%% file: main.tex
\documentclass[conference]{IEEEtran}
\IEEEoverridecommandlockouts

\usepackage{cite}
\usepackage{amsmath,amssymb,amsfonts}
\usepackage{graphicx}
\usepackage{textcomp}
\usepackage{xcolor}
\def\BibTeX{{\rm B\kern-.05em{\sc i\kern-.025em b}\kern-.08em
    T\kern-.1667em\lower.7ex\hbox{E}\kern-.125emX}}

\usepackage{booktabs}

\usepackage{algorithm} 
\usepackage{algpseudocode} 
\usepackage{svg}
\usepackage[style=base]{caption}
\usepackage[font=rm]{subcaption}
\usepackage{url}

\usepackage{pgfplots}

\def\markDot{\bullet}
\newcommand{\preset}[1]{^\markDot{}#1}
\newcommand{\postset}[1]{#1^\markDot{}}

\def\workcraft{\texttt{Workcraft}}
\def\petrify{\texttt{Petrify}}
\def\mpsat{\texttt{Mpsat}}
\def\reach{\texttt{REACH}}

\usepackage{ifthen}
\newboolean{anonymous}
\setboolean{anonymous}{false}

\begin{document}

\title
{
    Eventizing Traditionally Opaque Binary Neural Networks as 1-safe Petri net Models
}
 \author
 {
     \IEEEauthorblockN
     {
         Mohamed Tarraf, 
         Alex Chan, 
         Alex Yakovlev and 
         Rishad Shafik
     }
     \IEEEauthorblockA
     {
         Newcastle University, Newcastle upon Tyne, NE1 7RU, United Kingdom \\
         \{m.tarraf2, alex.chan, alex.yakovlev, rishad.shafik\}@newcastle.ac.uk
     }
 }
\maketitle

\begin{abstract}
Binary Neural Networks (BNNs) offer a low-complexity and energy-efficient alternative to traditional full-precision neural networks by constraining their weights and activations to binary values. However, their discrete, highly non-linear behavior makes them difficult to explain, validate and formally verify. As a result, BNNs remain largely opaque, limiting their suitability in safety-critical domains, where causal transparency and behavioral guarantees are essential. In this work, we introduce a Petri net (PN)–based framework that captures the BNN’s internal operations as event-driven processes. By ``eventizing'' their operations, we expose their causal relationships and dependencies for a fine-grained analysis of concurrency, ordering, and state evolution. Here, we construct modular PN blueprints for core BNN components including activation, gradient computation and weight updates, and compose them into a complete system-level model. We then validate the composed PN against a reference software-based BNN, verify it against reachability and structural checks to establish 1-safeness, deadlock-freeness, mutual exclusion and correct-by-construction causal sequencing, before we assess its scalability and complexity at segment, component, and system levels using the automated measurement tools in \workcraft{}. Overall, this framework enables causal introspection of transparent and event-driven BNNs that are amenable to formal reasoning and verification.
\end{abstract}

\begin{IEEEkeywords}
Binary neural network, Petri nets, machine learning, formal modeling, causal models.
\end{IEEEkeywords}

\section{Introduction}
Binary Neural Networks~(BNNs)~\cite{courbariaux2016binarizedneuralnetworkstraining} have emerged as a promising class of deep machine learning~(ML) models that offer a good balance between model accuracy and computational efficiency, where they have shown promising performance in energy-efficient on-device activity and stress recognition on wearable platforms~\cite{hosseini2019minimizing} and time-series health monitoring like arrhythmia classification~\cite{2023-ecai-pu_et_al-arrhythmia_classification_via_bnn}.
By constraining both weights and activations to binary values (i.e. +1 and -1), BNNs significantly reduce computational complexity and memory footprint for energy-efficient deployment on low-power processors and FPGA-based accelerators~\cite{Umuroglu_2017}, while also introducing a discrete, yet event-driven, structure through their bit-wise operations that is convenient for rigorous analysis~\cite{hubara2016quantized}.

However, BNNs remain \textit{opaque} as an ML model, as their internal decision-making processes involve computations that are both non-trivial and largely inaccessible~\cite{leblanc_Seeking_Interpretability_Explainability_Binary_Activated_Neural_Networks}. 
This opacity poses significant challenges in both explainability and analysis, especially for safety-critical applications like fault-tolerant satellite control and onboard system health monitoring, which require strict dependability and reliability, as any errors can lead to irreversible damage or even harm to users and operators~\cite{2021-ieee_comp-ashmore_et_al-why_explainability_is_important_for_ML_models_in_safety_critical_applications}.
Additionally, the discrete and highly non-linear transitions that are introduced by binarization can further exacerbate explainability, as any small changes to inputs or weights can cause abrupt changes to the BNN's activation, making gradient-based methods and conventional verification approaches difficult to apply~\cite{2021-cav-zhang_et_al-BDD4BNN}.
Still, while formal methods have traditionally focused on hardware and embedded systems like circuit and system-level verification~\cite{2018_springer_drechsler_formal-system-verification}, there is growing interest in applying similar principled techniques to ML models like BNNs to ensure correctness and reliability.

Given these challenges, it is crucial for engineers to have a clear insight into how these models make their decisions by validating their behavior under edge cases, reasoning about any fault propagation, and establishing guarantees of correctness, robustness and resilience.
A natural way to address the BNN's opacity is through \textit{causality}, where one can describe the relationship between events and establish what events directly influence or determine the occurrence of another event. 
For example, in safety-critical systems, understanding causality is essential to help predict system behavior under anomalies, trace fault propagation and ensure dependable operation.
This causal perspective of BNNs allows engineers to understand how a specific combination of inputs, weight configurations, and internal activations contribute to its output by providing a mechanistic view that goes beyond black-box approximations. 

Traditional methods, such as SHAP~(SHapley Additive exPlanations)~\cite{2017-neurips-lundberg_scott_lee-SHAP} and LIME~(Local Interpretable Model-agnostic Explanations)~\cite{2016-kdd-ribeiro_singh_guestrin-LIME}, only offer post-hoc correlation driven approximations, while verification approaches like satisfiability modulo theories (SMT)-based reasoning~\cite{2017-CAV-springer-katz_et_al-SMT_verification} and convex relaxation~\cite{2018-icml-wong_et_al-convex_relaxation} generally struggle to capture the fine-grained event dependencies in the BNN's computation. 
By explicitly capturing these causal dependencies, engineers can analyze, validate and formally verify the network's behavior for guarantees of reliability, robustness, and correctness. 

This need for explicit causal reasoning motivates the use of Petri nets (PNs)~\cite{1989-murata-ieee_journal-petrinets} as their token-based semantics and discrete event-based structure are designed to easily capture causality, concurrency, and state transitions.
Since every binary activation, thresholding step, and weight update in BNNs can be represented as a PN transition, the complete operational semantics of BNNs can be captured to trace dependencies, validate behavior, and verify structural and behavioral properties. 
Moreover, extensions like generalized stochastic PNs~\cite{2007-springer-sfm-balbo-generalized_stochastic_petri_nets} further demonstrate the suitability of PNs for encoding complex system dynamics in contexts such as probabilistic model checking~\cite{2021-dsn-khan_et_al-figaro_featuring_petri_nets}.
In fact, recent work has also shown how PNs can be used 
as rigorous executable models for learning systems by modeling Tsetlin Machines~(TMs)~\cite{2018-granmo-tsetlin} as 1-safe PNs for event-triggered simulation, validation, and verification~\cite{2024-chan_wheeldon_shafik_yakovlev-event_driven_tsetlin_machines_using_petri_nets}.
This particular work primarily focused on TMs that are already naturally explainable through their \textit{white-box}-like architectures, and are automata-based learners that do not rely on differentiable objectives and gradients, which means that this approach does not yet support ML models that contain continuous or approximated real-valued weights.

In this paper, we present a PN-based framework that systematically captures the BNN's inference and training dynamics, where we explicitly model the causal structure of BNN's computation to a fine-granular level revealing its concurrency and state evolution from data loading to calculating new weights.
This provides a mechanistic view of how the BNN's internal operations work, while also enabling it for rigorous behavioral validation, correctness checking, and behavioral and structural verification that were previously infeasible for these models.
To support this, we use the \workcraft{}~\cite{2006-workcraft-website} toolset that supports PN construction, interactive event-level simulation, and automated verification through its \petrify{}~\cite{1997-cortadella-ieice-petrify} and \mpsat{}~\cite{2004-khomenko-acsd-mpsat} backend, which allows us to examine causal dependencies, validate PN-based BNN executions, and formally verify key structural properties.
\workcraft{} has also recently seen use by industrial practitioners~\cite{2025-async-sokolov_khomenko_sautto-easy_async_design_AFSM_workcraft} for developing energy-efficient batteries for smart phones~\cite{2020-sokolov-tcad-a4a}. 
Together, this framework and tooling pipeline provide a principled foundation for bringing high-performance ML models into settings that demand dependability and rigorous analysis.

Thus, the contributions of this paper include:
\begin{itemize}
    \item A systematic methodology for modeling event-driven BNNs by using blueprint-like PN segments that capture their inference and learning processes (Section~\ref{sect:Eventizing the Binary Neural Network}).
    \item Automated property checking of the BNN–PN model using \workcraft{}'s \mpsat{} backend to verify key structural and behavioral properties like 1-safeness and deadlock-freeness with event-by-event analysis of how weights are updated via causal dependencies 
    (Section~\ref{sect:verification}). 
    \item Behavioral validation of the BNN-based PN model, where PN-based execution is compared against a reference BNN and analytical data is collected via an instrument using \workcraft{}'s simulation tool (Section~\ref{sect:validating_bnn_pn_against_reference}).
    \item A quantitative analysis of structural scalability by assessing model size (Section~\ref{sect:model-size-evaluation}) and complexity (Section~\ref{sect:estimated-complexity}) at segment, component, and full-system levels.
\end{itemize}

The remainder of the paper goes as follows. 
Section~\ref{section:preliminary} covers the necessary background in BNNs, PNs and \workcraft{}.
Section~\ref{sect:Eventizing the Binary Neural Network} illustrates how the BNN's operations are captured using PNs. 
Section~\ref{sect:verification} describes how PN-based BNNs are verified on key PN properties.
Section~\ref{sect:validation-and-experiment-results} highlights our experimental results of the PN-based BNNs, with an evaluation on its model size and model complexity.
Section~\ref{sect:conclusion} concludes this paper and underlines potential future works.

\section{Preliminaries}\label{section:preliminary}

In this section, we cover the necessary background in BNNs and PNs, as well as the \workcraft{}~\cite{2006-workcraft-website} toolset, to help understand the work in the later sections.

\subsection{Binary Neural Networks}\label{sect:preliminary:BNNs}
A BNN is a class of DNN in which both weights and activations are discretized to binary values, typically +1 and -1~\cite{courbariaux2016binarizedneuralnetworkstraining}. As with conventional neural networks, BNNs operate in two main phases: \textit{inference}, where inputs are propagated through the network to produce an output, and \textit{training}, where parameters are updated to reduce the discrepancy between the actual and desired outputs. A BNN follows the standard feedforward architecture consisting of an input layer, one or more hidden layers, and an output layer. Each neuron in a hidden layer receives one weight per input feature; these weights are initially assigned real valued random floats and subsequently binarized using the \textit{Sign} function in equation~\eqref{eq:Sign}.
\begin{equation}
    x_b = \operatorname{Sign}(x) =
    \begin{cases}
        +1 & \text{if } x \geq 0,\\
        -1 & \text{if } x < 0
    \end{cases}
    \label{eq:Sign}
\end{equation}

During inference, the input vector is processed sequentially through weighted sums and activation functions. Each neuron computes a pre-activation value by taking the dot product of the input features with their binary weights. This value is then passed through the Sign function (or, in some BNN variants, a continuous surrogate such as TanH) to obtain a binary activation. In the output layer, the activations of all hidden neurons are aggregated and passed through the output activation function to compute the network’s prediction.

Training begins with the initialization of the learning rate and full-precision (real valued) weights. After a forward pass, the network output is evaluated using a loss function that quantifies the error relative to the expected output. The loss is then differentiated to obtain gradients for weight updates. However, because the \textit{Sign} function is non-differentiable, BNNs use a Straight-Through Estimator (STE) to approximate its gradient~\cite{bengio2013estimatingpropagatinggradientsstochastic}~\cite{yin2019understandingstraightthroughestimatortraining}, equation~\eqref{eq:STE}.
\begin{equation}
    \frac{\partial x_b}{\partial x} \approx \text{STE} =
    \begin{cases}
        1 & \text{if } |x| \leq 1, \\
        0 & \text{otherwise}
    \end{cases}
    \label{eq:STE}
\end{equation}

Using the STE, gradients are first computed with respect to the binary weights and then mapped back to the underlying full-precision weights. Finally, the full-precision weights are updated according to the chosen optimization rule, after which they are binarized again for the next forward propagation step.

\begin{figure}[]
    \centering
    \begin{subfigure}[b]{0.17\textwidth}
        \centering
        \includegraphics[width=\linewidth]{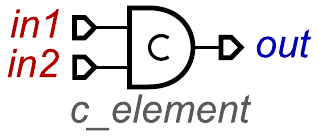}
        \hspace{50px}
        \caption{}
        \label{fig:c-element:gate}
    \end{subfigure}
    \begin{subfigure}[b]{0.31\textwidth}
        \centering
        \includegraphics[width=\linewidth]{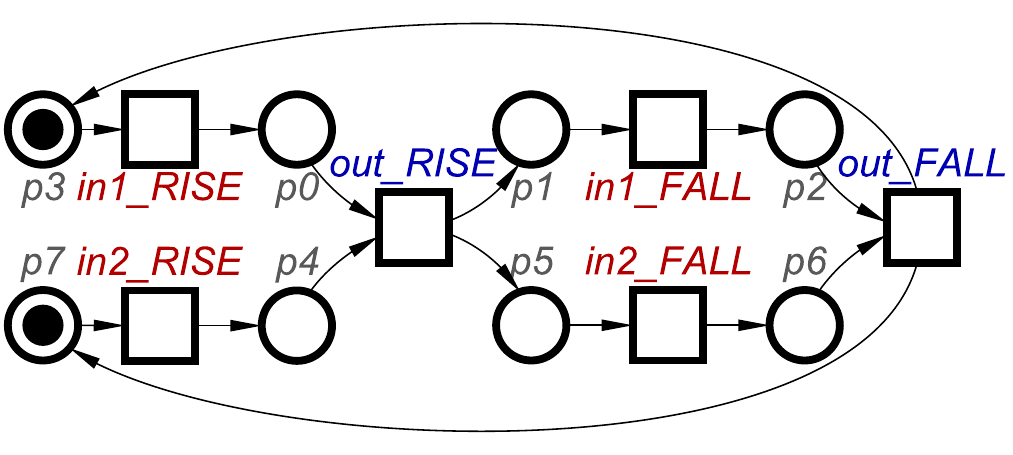}
        \caption{}
        \label{fig:c-element:petrinet}
    \end{subfigure}
    \caption{C-element Specification. (a) Logic gate. (b) PN model.}
    \label{fig:c-element}    
\end{figure}

\subsection{Petri nets}\label{sect:preliminary:petri-nets}
PNs~\cite{1989-murata-ieee_journal-petrinets} are a simple, yet powerful, mathematical model that can be used to specify and analyze many systems like distributed systems and asynchronous circuits~\cite{1985-yakovlev-petrinets_workshop-stg}.
They can be formally defined as a tuple $PN = (P, T, F, M, l)$ where:
\begin{itemize}
    \item $P$ is the finite set of places.
    \item $T$ is the finite set of transitions, such that $T \cap P \neq \emptyset$.
    \item $F \subseteq (P \times T) \cup (T \times P)$ is the set of arcs from places to transitions and transitions to places. 
    \item $M$ is a marking of $PN$ where $M: P \rightarrow \{0, 1, 2, ...\}$ such that $M_0$ is the initial marking.
    \item $l: T \rightarrow \Sigma$ is a labeling function of transitions, where $\Sigma$ is the language alphabet.
\end{itemize}

Graphically, the PN's places are represented as circles, transitions are represented as squares, arcs are represented as arrows, and tokens are represented as black dots (or a number) inside places, as shown in Fig.~\ref{fig:c-element} where the design of a two input C-element gate is 
depicted.

Here, the notation for preset and postset of places in PNs can be written as \(\preset{x}=\{\,p\in P \mid (p,x)\in F\,\}\) and \(\postset{x}=\{\,p\in P \mid (x,p)\in F\,\}\) for any \(x\in P\cup T\) respectively. 
The incidence matrix \(C\in\mathbb{Z}^{|P|\times|T|}\) has entries \(C(p,t)=W(t,p)-W(p,t)\), where \(W(\cdot,\cdot)\) is the arc weight that equals \(1\) in this paper. 

A transition $t \in T$ is said to be enabled at marking $M$ when every input place holds at least one token, $\forall p\in\preset{t}:\; M(p)\ge 1$.
When \(t\) is fired, the marking changes to $M' = M + C(:,t)$ where \(C(:,t)\) is the \(t\)-th column of \(C\). 

\subsection{Workcraft Toolset}\label{sect:preliminary:WorkcraftToolset}
\workcraft~\cite{2006-workcraft-website} is a visual design framework that offers support for the automation of interpreted graph-based models, including PNs, Finite State Machines~~\cite{1962-gill-introduction_to_fsm}, and Signal Transition Graphs~\cite{1985-yakovlev-petrinets_workshop-stg}. It provides an intuitive graphical front-end for constructing, editing, and simulating these models, complemented by a rich toolchain, notably \petrify{}~\cite{1997-cortadella-ieice-petrify} and \mpsat{}~\cite{2004-khomenko-acsd-mpsat} for verification and circuit synthesis.

A key design feature employed in this work is the use of proxy places (i.e. proxies). A proxy is a place that has been visually ``collapsed'', where its incoming arcs from preset transitions and outgoing arcs to postset transitions are hidden and replaced with a compact textual label representing the place. 
Proxies are only visually detached, as they do not alter the underlying semantics or behavior of the PN, since the original place is still present in the model. 
This feature helps enhance readability and visual analysis by reducing arc clutter, preventing overlapping connections, and allowing specific places, such as those marking entry into a critical section, to be tracked without navigating the whole PN. 

\section{Eventizing Binary Neural Networks}\label{sect:Eventizing the Binary Neural Network}
\begin{figure} 
    \centering
    \includegraphics[width=0.75\linewidth]{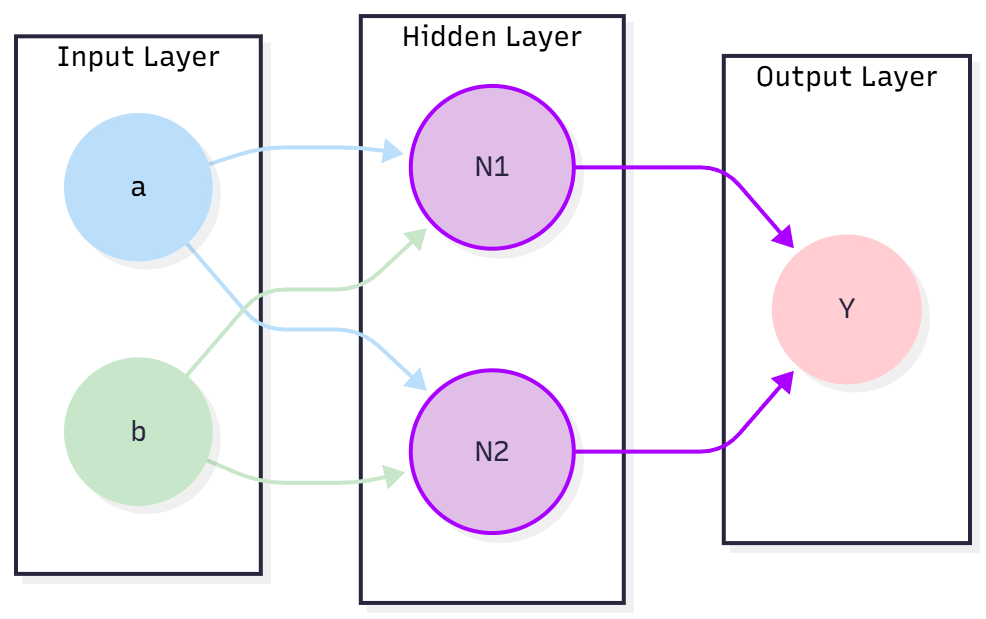}
    \caption{Architecture of BNN model performing 2-input XOR operation - used as an illustrative example.}
    \label{fig:Structure}    
\end{figure}

In this section, we present the methodology used to model a BNN as an event-driven system using PNs. For illustration purposes, the BNN architecture considered here consists of an input layer with two neurons, a hidden layer with two neurons, and a single-neuron output layer, as shown in Fig.~\ref{fig:Structure}.
Note that the impact on the size of the BNN PN from scaling up the number of neurons or increased complexity of datasets is later shown in Section~\ref{sect:estimated-complexity}.
For clarity, the model excludes bias terms. We focus on modeling both inference and training hierarchically: each operation is first represented as a small PN segment, these segments are then composed into inference and training components, and finally integrated into a complete BNN-level model. The network is trained to learn the XOR pattern using the dataset \{00, 01, 10, 11\} with corresponding expected outputs \{-1, 1, 1, -1\}. Each hidden neuron includes two input–hidden weights and one hidden–output weight.

\subsection{Design of Inference Segments }\label{sect:Eventizing the Binary Neural Network:Design of Inference Segments}
\begin{figure} 
    \centering
    \includegraphics[width=0.9\linewidth]{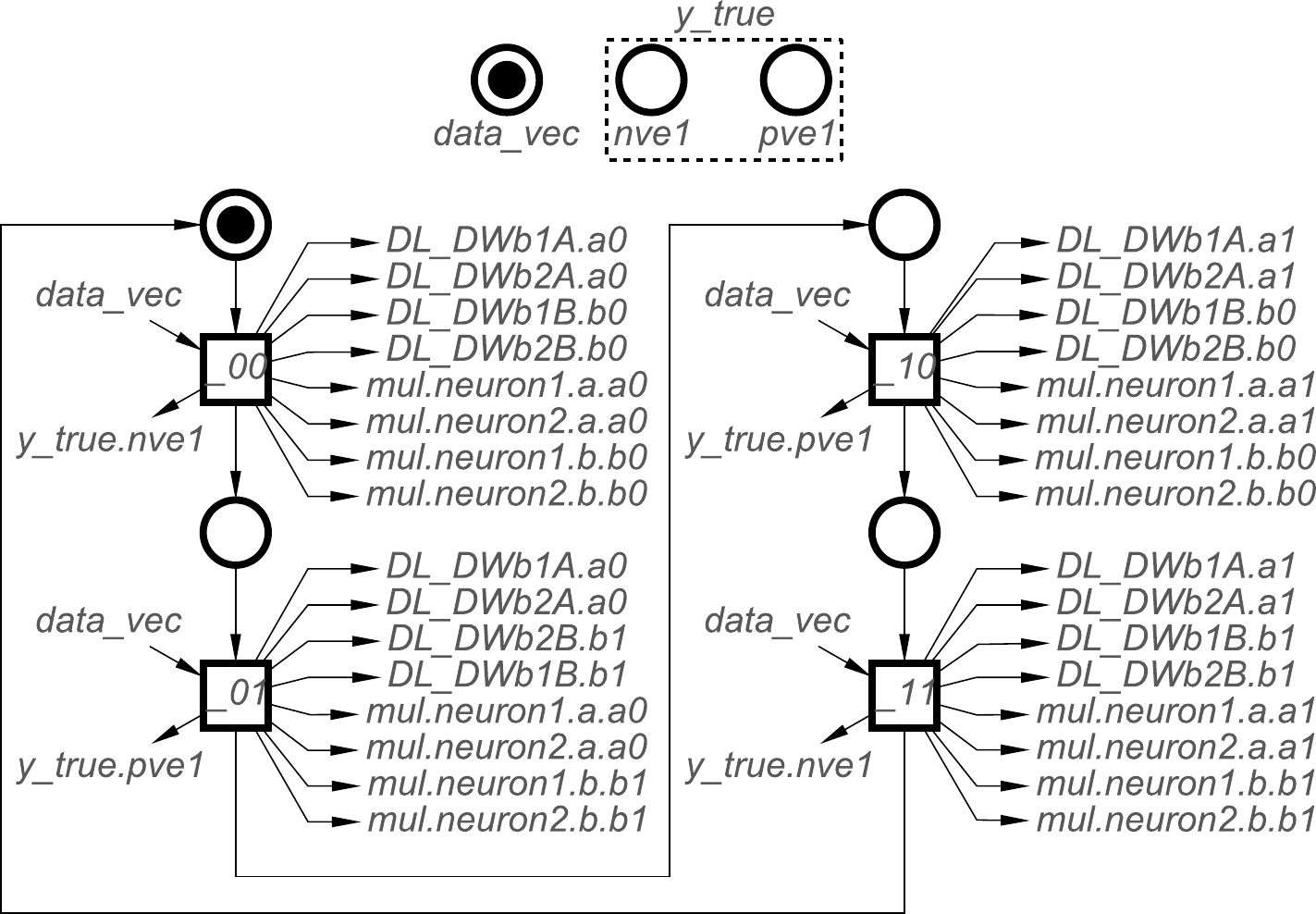}
    \caption{PN segment of data point loading with transitions representing data points, places for data vector initialization, and places for expected output.}
    \label{fig:Inputs}    
\end{figure}

The loading of input data points is modeled using four transitions labeled 00, 01, 10, and 11, where we connect them to and from places that determine the order of how the data points are loaded (00 to 11), when the next data point is loaded (when there is a token in \textit{data\_vec}), and to a place that represents the expected value, this can be seen in Fig. \ref{fig:Inputs}. Connections to other components are represented using proxies.

\begin{figure}
    \centering
    \includegraphics[width=0.9\linewidth]{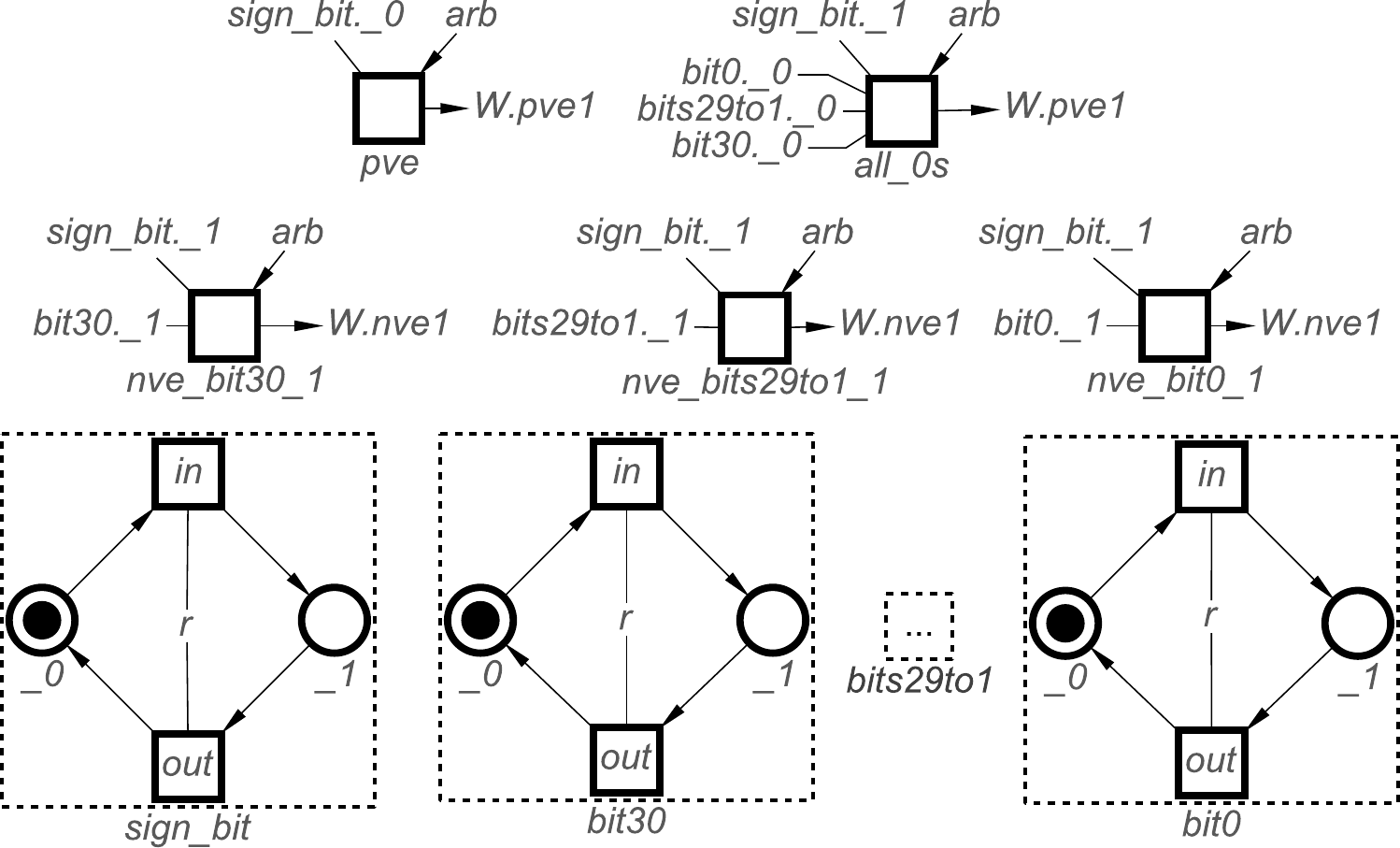}
    \caption{PN segment of initialization and binarization of real valued weights.}
    \label{fig:WeightBin}
\end{figure}

Weight initialization and binarization are combined within a single PN segment. Each bit of a weight is represented as a two-place buffer (value 0 or 1) controlled by transitions \textit{in} and \textit{out} that toggle the stored value. 
Fig.~\ref{fig:WeightBin} illustrates the binarization of the 32 bit IEEE-754 floating point number, where bit 31 represents the sign bit, bits 30 to 23 represents the exponent, and bits 22 to 0 represents the mantissa. All buffer transitions use read arcs from place \textit{r}, ensuring they can fire only while \textit{r} holds a token. After the transition \textit{set\_weights} fires, the token moves from \textit{r} to place \textit{arb}, preventing any further modification of the bits. Binarization proceeds as follows. If the sign bit holds 0, transition \textit{pve} fires and the weight is binarized to +1 regardless of the value bits. If the sign bit holds 1 and all value bits are 0, transition \textit{all\_0s} fires, also producing +1. Otherwise, if the sign bit is 1 and any value bit is 1, one of several equivalent transitions may fire, binarizing the weight to –1. Multiple transitions may be enabled, but the choice does not affect the final output. Place \textit{arb} exists to ensure that only one binarization transition fires, keeping model safe. 

\begin{figure}
    \centering
    \includegraphics[width=\linewidth]{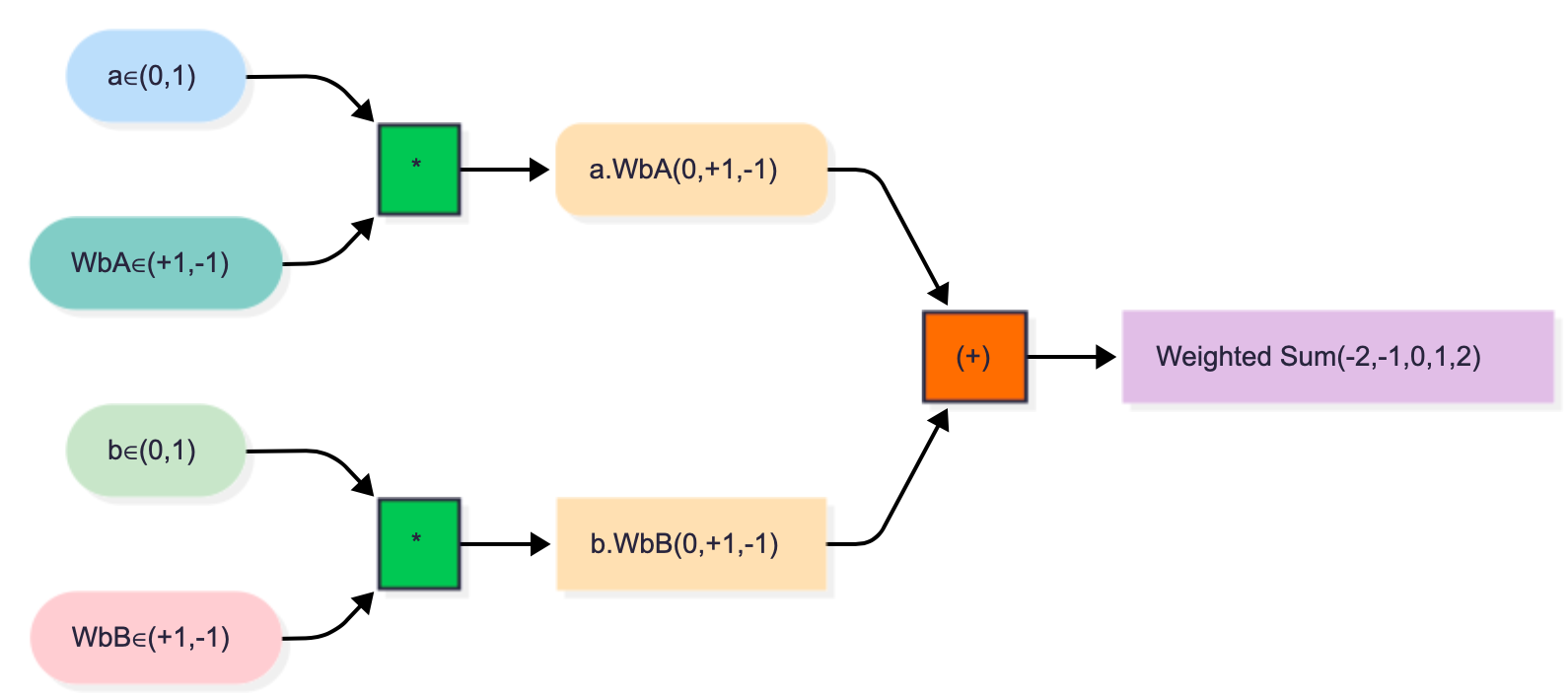}
    \caption{Sequence of operations demonstrating weighted sums for pre-activation of a neuron.}
    \label{fig:PreAc}
\end{figure}

The pre-activation value for each neuron is computed by multiplying each input feature with its corresponding binary weight. The outputs of those operations are then summed to get a weighted sum. This can be better visualized in Fig.~\ref{fig:PreAc}. With PNs, all reachable products are modeled as explicit PN paths for each feature and weight and the same applies to each reachable sum of their outputs.

\begin{figure} 
    \centering
    \begin{subfigure}[b]{0.2\textwidth}
        \centering
        \includegraphics[width=0.9\linewidth]{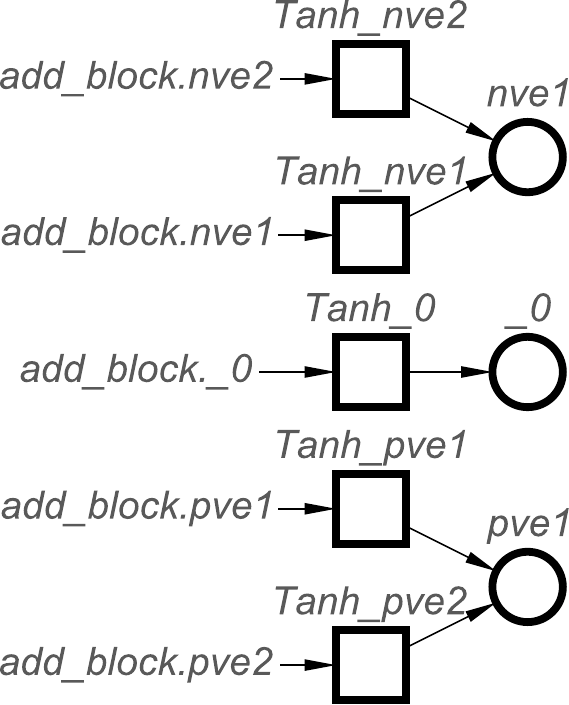} 
        \caption{}
        \label{fig:Activation:TanH}
    \end{subfigure}
    \hspace{5px}
    \begin{subfigure}[b]{0.2\textwidth}
        \centering
        \includegraphics[width=0.75\linewidth]{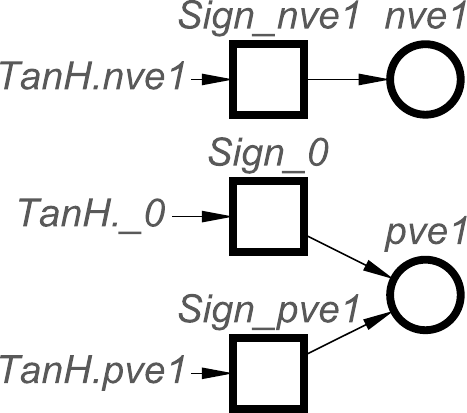}
        \caption{}
        \label{fig:Activation:Sign}
    \end{subfigure}
    \caption{PN segment for activation. (a) Hard Tan function. (b) Sign function.}
    \label{fig:Activation}    
\end{figure}

\begin{figure}
    \centering
    \includegraphics[width=0.85\linewidth]{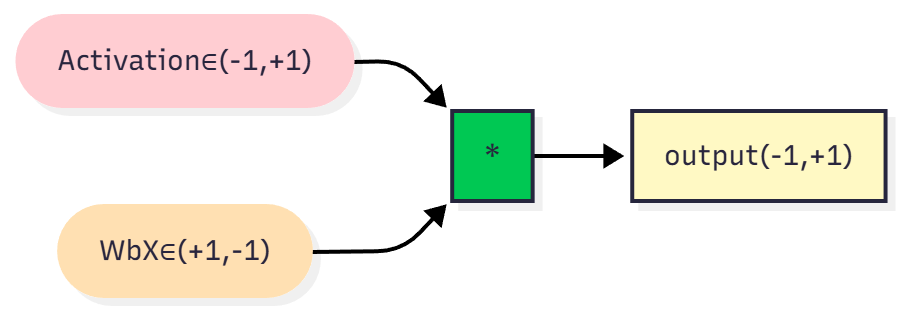}
    \caption{Neuron output operations using activations and binary output weights.}
    \label{fig:NeuronOut}
\end{figure}

BNN activations are commonly computed using the \textit{Sign} function, though some variants instead apply \textit{TanH}, which produces continuous values between +1 and -1. In our PN design, we model both functions to enable verification of their behavior. The pre-activation value is first passed through the \textit{TanH} function (Fig.~\ref{fig:Activation:TanH}), which compresses the output from Fig.~\ref{fig:PreAc}. Then it passes through the \textit{Sign} function (Fig.~\ref{fig:Activation:Sign}) to produce the final discrete activation. The neuron’s output is then obtained by multiplying the activation by the binary hidden-to-output weight, as shown in Fig.~\ref{fig:NeuronOut}. This sequence of operations (and PN segments) is unanimous across all hidden neurons.

\begin{figure}
    \centering
    \includegraphics[width=\linewidth]{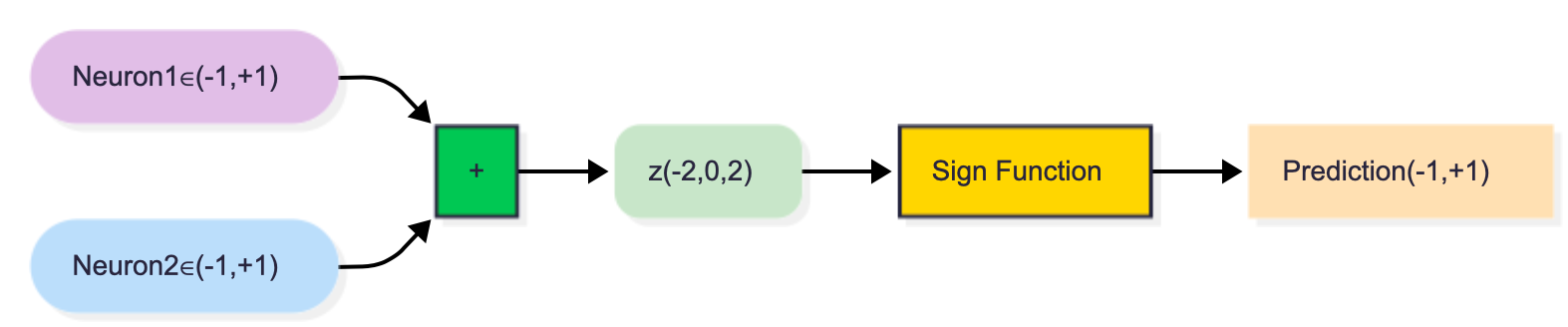}
    \caption{Operations showcasing sum of neuron outputs and the final prediction using Sign function.}
    \label{fig:Output}
\end{figure}

The outputs of all neurons are then summed,in the same fashion as pre-activation, and passed through another \textit{Sign} function segment to produce the final prediction of the inference stage, +1 or –1, in line with the expected output format, Fig.~\ref{fig:Output}. This completes the inference pipeline of the BNN.

\subsection{Design of Training Segments}
Training begins by computing the loss from the inference stage, that is, how far the predicted output deviates from the expected output. We use Hinge Loss~\cite{Rennie2005LossFF}, defined in equation~\eqref{eq:Hinge}, where z is the summed neuron output and y is the expected output from Fig.~\ref{fig:Inputs}. 

Hinge Loss, originally introduced in the context of Support Vector Machines (SVMS)~\cite{Boser.et.al.}, is widely used in BNN training, in both its raw and squared variants, due to its strong performance~\cite{courbariaux2016binarizedneuralnetworkstraining}, as well as it matching the output structure of a BNN (+1 and -1). Its computational simplicity makes it straightforward to model with PNs and also provides a simple derivative, shown in equation~\eqref{eq:DL}.

\begin{equation}
    L=\max(0, 1-y \cdot z)
    \label{eq:Hinge}
\end{equation}
\begin{equation}
    \frac{\partial L}{\partial z} =
    \begin{cases}
        -y & \text{if } y \cdot z < 1, \\
        0 & \text{if } y \cdot z \geq 1
    \end{cases}
    \label{eq:DL}
\end{equation}

\begin{figure} 
    \centering
    \begin{subfigure}[b]{0.25\textwidth}
        \centering
        \includegraphics[width=\linewidth]{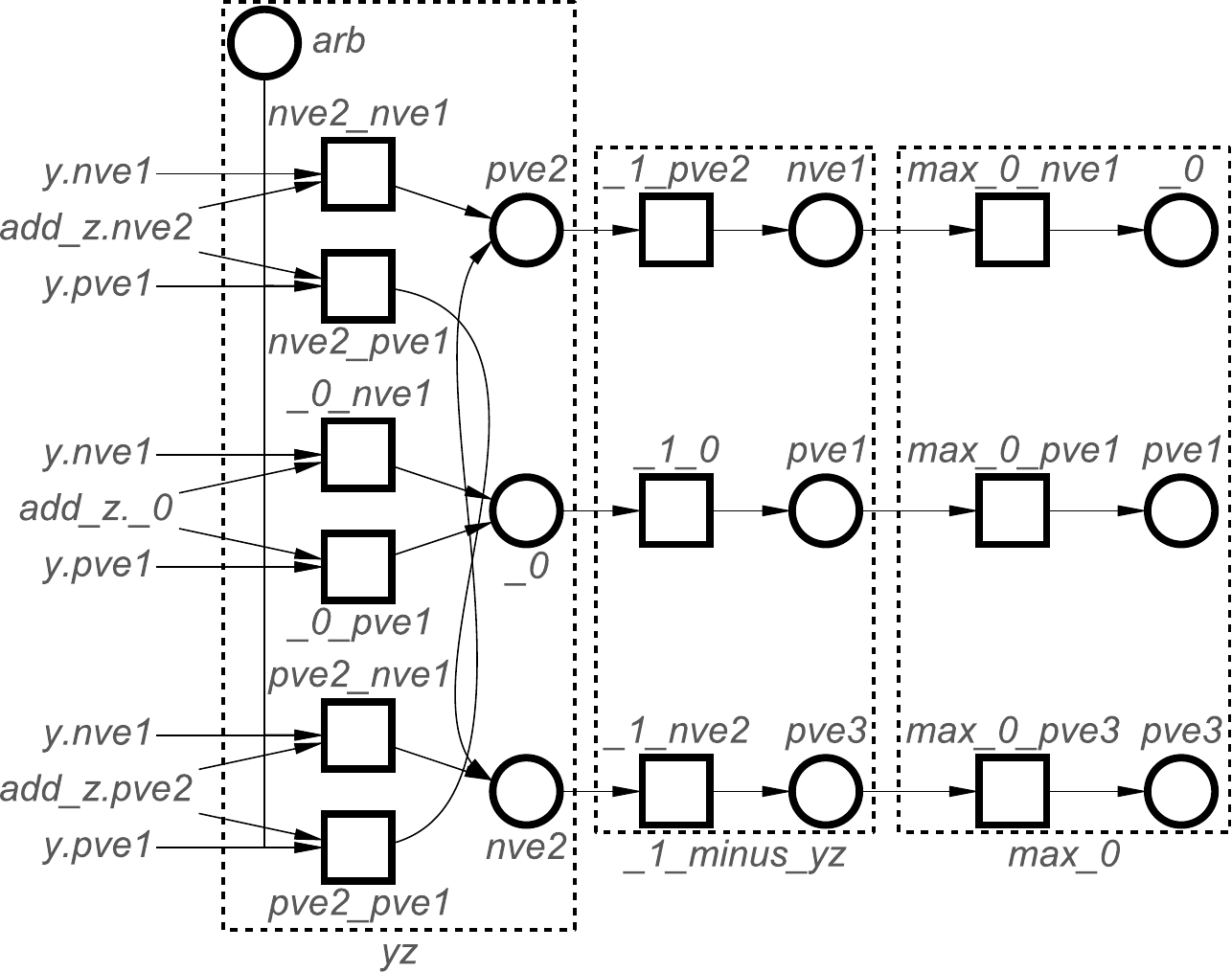} 
        \caption{}
        \label{fig:Loss:Hinge Loss}
    \end{subfigure}
    \begin{subfigure}[b]{0.23\textwidth}
        \centering
        \includegraphics[width=\linewidth]{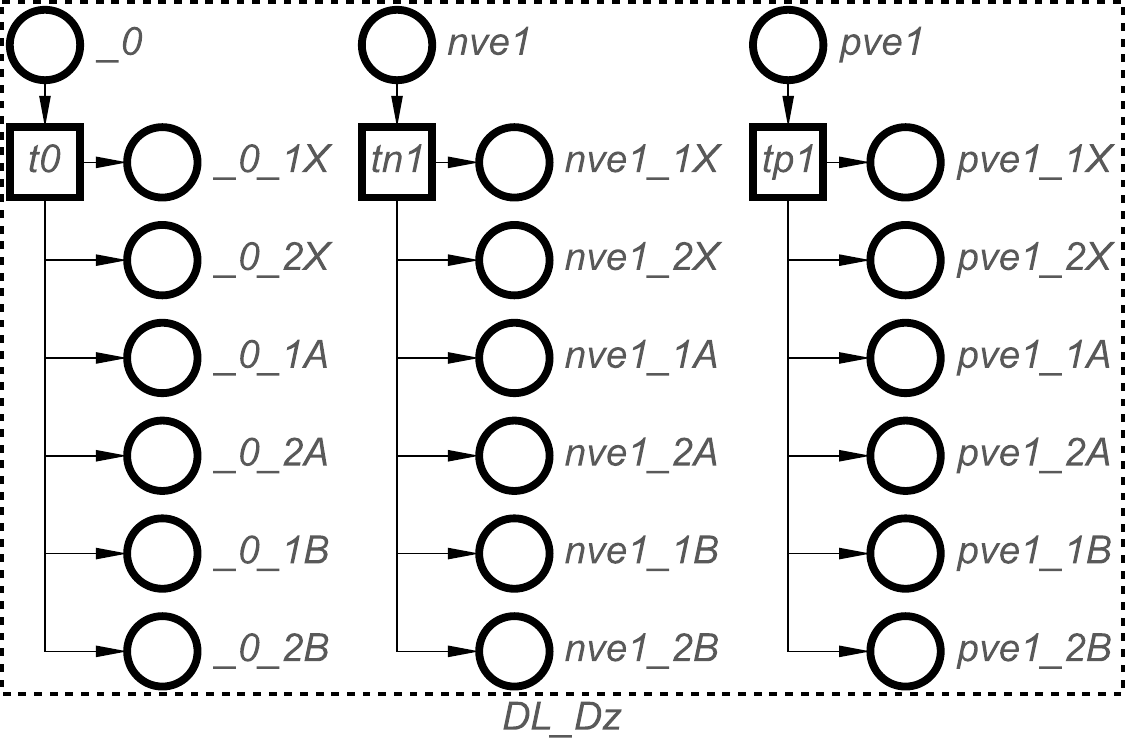}
        \caption{}
        \label{fig:Loss:Derivative}
    \end{subfigure}
    \caption{PN segment for Output layer. (a) Hinge Loss function. (b) Hinge Loss derivative.}
    \label{fig:Loss}    
\end{figure}

The PN segment for Hinge Loss, shown in Fig.~\ref{fig:Loss:Hinge Loss}, is divided into three parts.
\begin{enumerate}
    \item \textit{Multiplication}: It first computes the product of z (sum of neuron outputs) and y (expected output)
    \item \textit{Subtraction}: It then subtracts this value from 1.
    \item \textit{Clipping}: Finally, it selects the maximum of 0 and the result of step two, yielding the loss score.
\end{enumerate}
A loss of 0 indicates a correct classification with full confidence; a loss of 1 reflects either a misclassification or a correct classification with low confidence; and a loss of 3 indicates a misclassification. The derivative of the loss is determined during the first step. Arcs route this derivative to the appropriate output place in Fig.~\ref{fig:Loss:Derivative}, where the token is then forked into six places for computing gradients with respect to each weight.

\begin{figure}
    \centering
    \includegraphics[width=\linewidth]{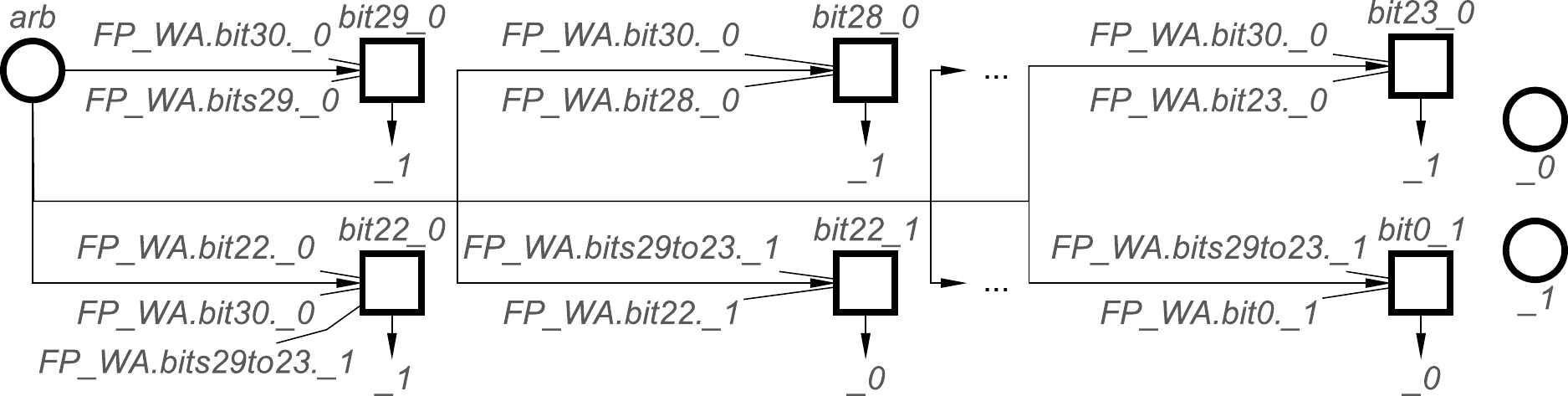}
    \caption{PN segment showcasing the BNN's STE}
    \label{fig:STE}
\end{figure}

The STE computation (equation~\eqref{eq:STE}) is illustrated in Fig.~\ref{fig:STE}. Multiple transitions capture possible weight combinations while keeping the total number of transitions minimal for scalability. This design allows values to be mapped to 1 or clipped to 0 as required. As with weight binarization in Fig.~\ref{fig:WeightBin}, several transitions may be enabled simultaneously, but all produce identical outputs.

We now compute and backpropagate gradients according to the update rule of the chosen optimizer. In this work, we use Stochastic Gradient Descent (SGD)~\cite{Robbins1951ASA}. SGD updates each weight by moving it in the direction opposite to the gradient of the loss. We incorporate SGD into our model because its update rule is structurally simple and therefore well suited for PN implementation. In contrast, more sophisticated BNN optimizers such as ADAM~\cite{kingma2017adammethodstochasticoptimization} rely on moving averages and additional state variables, which can significantly complicate their representation within a PN framework.

\begin{figure}
    \centering
    \includegraphics[width=\linewidth]{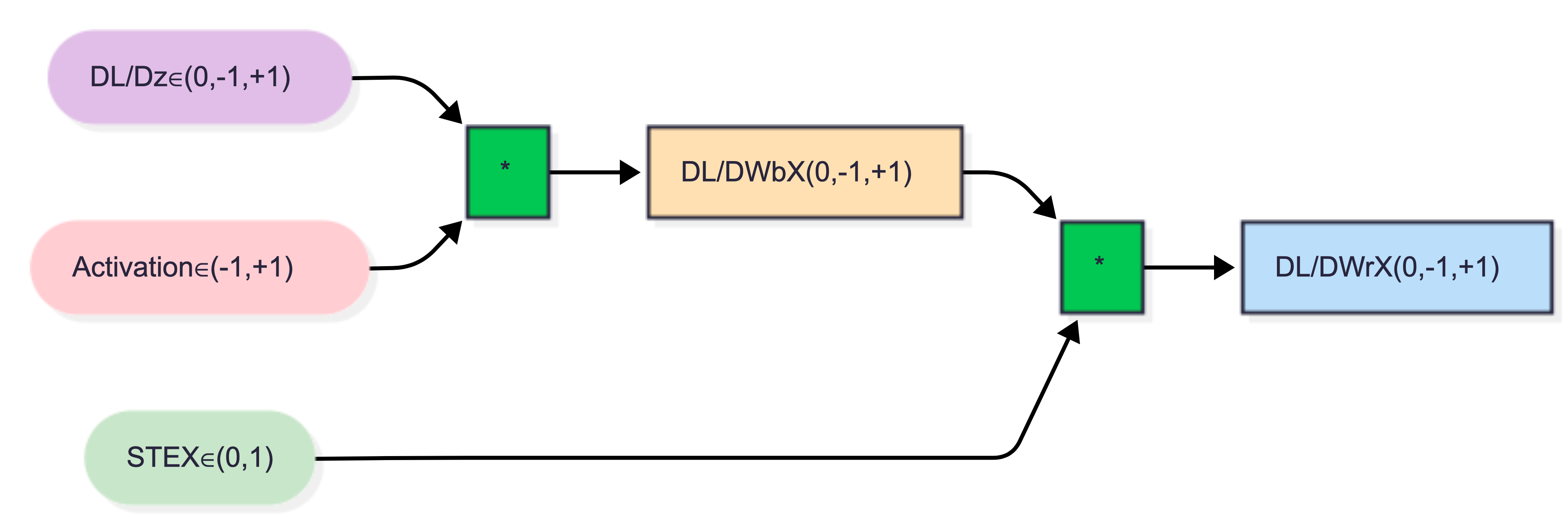}
    \caption{Sequence of operations for gradient computations.}
    \label{fig:Grad}
\end{figure}

The gradient with respect to the binary input–hidden weights follows equation~\eqref{eq:GradWbih}, and the gradient with respect to the binary hidden–output weight follows equation~\eqref{eq:GradWbho}. The gradients with respect to the real-valued weights are computed using equation~\eqref{eq:GradWr} and are modeled in the same way as the previous operation. These equations breakdown the inputs going into each segment and so using that information, we compose PN segments accordingly. We can visualize the sequence of operations for the gradient w.r.t the hidden-output weights in Fig.~\ref{fig:Grad}, but the same concept applies for the input-hidden weights.

\begin{equation}
    \frac{\partial L}{\partial W_bA} = \frac{\partial L}{\partial z} \cdot \frac{\partial z}{\partial x} \cdot \frac{\partial x}{\partial s} \cdot \frac{\partial s}{\partial W_bA}= \frac{\partial L}{\partial z} \cdot W_bx \cdot 1 \cdot a
    \label{eq:GradWbih}
\end{equation}
\begin{equation}
    \frac{\partial L}{\partial W_bx} = \frac{\partial L}{\partial z} \cdot \frac{\partial z}{\partial W_bx}= \frac{\partial L}{\partial z} \cdot x
    \label{eq:GradWbho}
\end{equation}
\begin{equation}
    \frac{\partial L}{\partial W_r} = \frac{\partial L}{\partial W_b} \cdot \frac{\partial W_b}{\partial W_r}=\frac{\partial L}{\partial W_b} \cdot \text{STE}
    \label{eq:GradWr}
\end{equation}

\begin{figure} 
    \centering
    \includegraphics[width=0.85\linewidth]{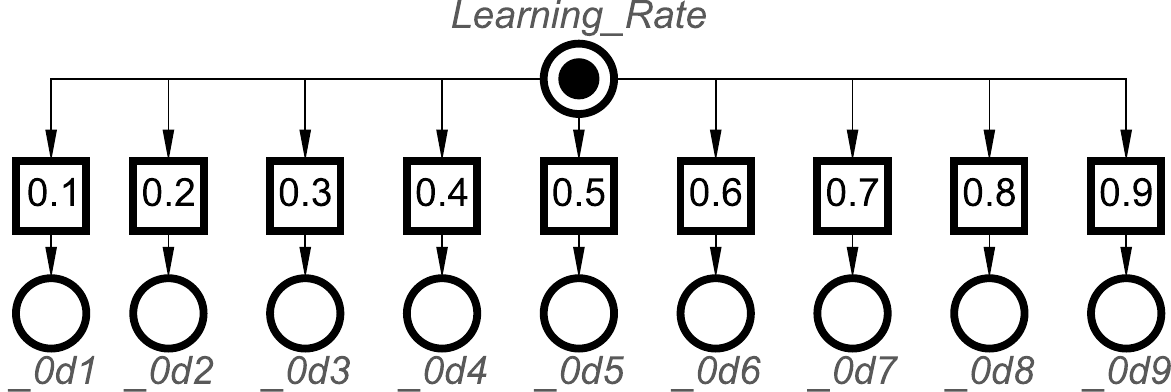}
    \caption{PN segment displaying choice of learning rates.}
    \label{fig:LR}    
\end{figure}

\begin{figure}
    \centering
    \includegraphics[width=\linewidth]{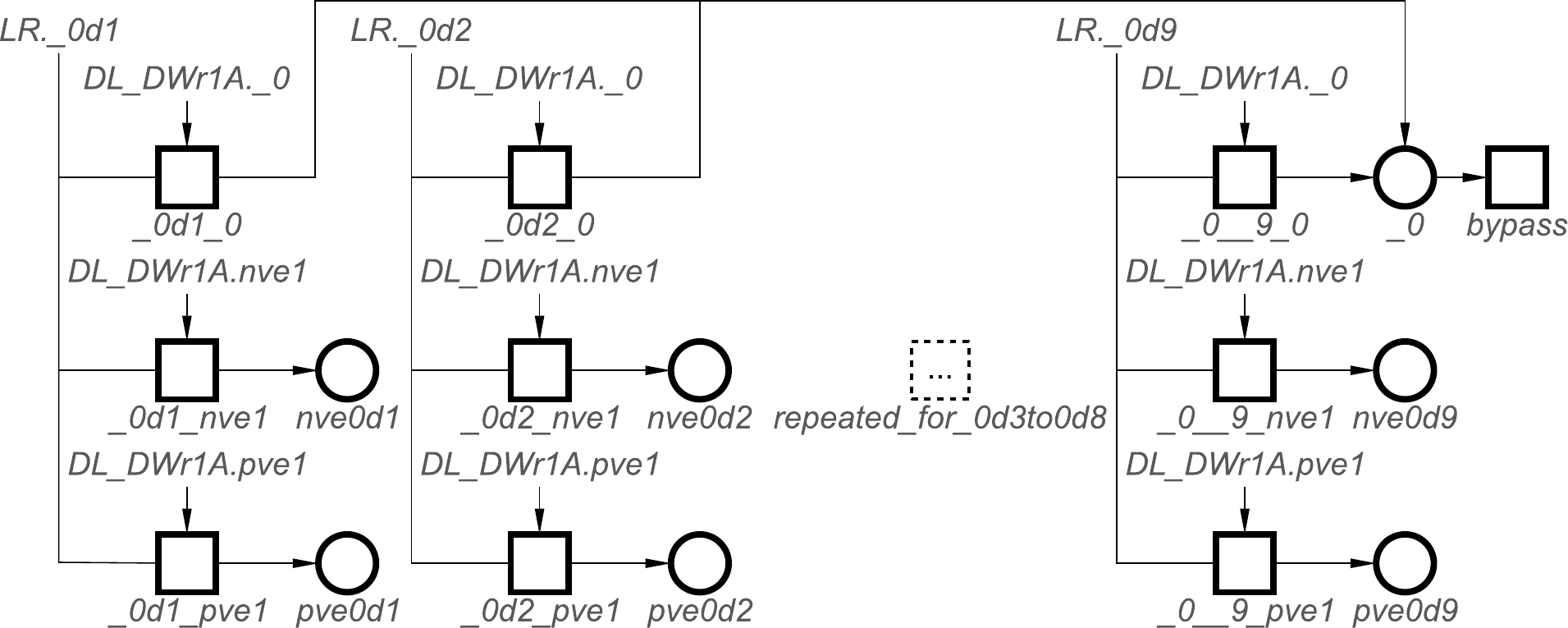}
    \caption{PN segment displaying product of learning rate and gradient.}
    \label{fig:LRGradWr}
\end{figure}

The SGD update rule is straightforward: each weight is adjusted by subtracting the product of the learning rate and its corresponding gradient (equation~\eqref{eq:LRGradWr}). Figure~\ref{fig:LR} shows the learning rate initialization, where nine possible learning rates, from 0.1 to 0.9, are provided. These learning rate places are connected to the computation segment in Fig.~\ref{fig:LRGradWr}, which multiplies the selected learning rate with the gradient, providing a negative and positive output for each learning rate value and 0. If the product of the learning rate is 0, then the subtraction then the weight update component is bypassed as the weight value does not change. Because the learning rate places use read-arc connections to the multiplication component, the learning rate needs to be fired only once for the entire system simulation.

\begin{equation}
    W_\text{r new} = W_r-\eta \cdot \frac{\partial L}{\partial W_r}
    \label{eq:LRGradWr}
\end{equation}

Updating the weights is conceptually simple (equation~\eqref{eq:LRGradWr}); however, because we use floating-point representations, performing binary subtraction involves several non-trivial steps:
\begin{itemize}
    \item The weight $W$ is already in floating-point binary format, but $J$ ($=\eta \cdot {\partial L}/{\partial W_r}$) is initially in decimal form and must be converted to floating-point binary.
    \item The new sign bit is determined using a truth-table–style comparison of the original sign bits and the relative magnitudes of $W$ and $J$
    \item The exponents of $W$ and $J$ are aligned by right-shifting the mantissa.
    \item The aligned mantissas are then added or subtracted, depending on whether an addition or subtraction magnitude is needed.
    \item The resulting mantissa is normalized by shifting left until it fits the normalized floating-point format.
    \item The new exponent is computed by subtracting the number of normalization shifts from the value 127.
\end{itemize}
This sequence mirrors the standard IEEE-754 floating-point subtraction procedure.

Converting J from decimal to binary floating-point format is done by connecting the outputs of Fig.~\ref{fig:LRGradWr} to their corresponding transitions. Each transition maps the computed value to the appropriate bit-position place (0 or 1 for each of the 32 bits), producing the floating-point representation required for subsequent operations.

\begin{figure} 
    \centering
    \begin{subfigure}[b]{0.48\textwidth}
        \centering
        \includegraphics[width=\linewidth]{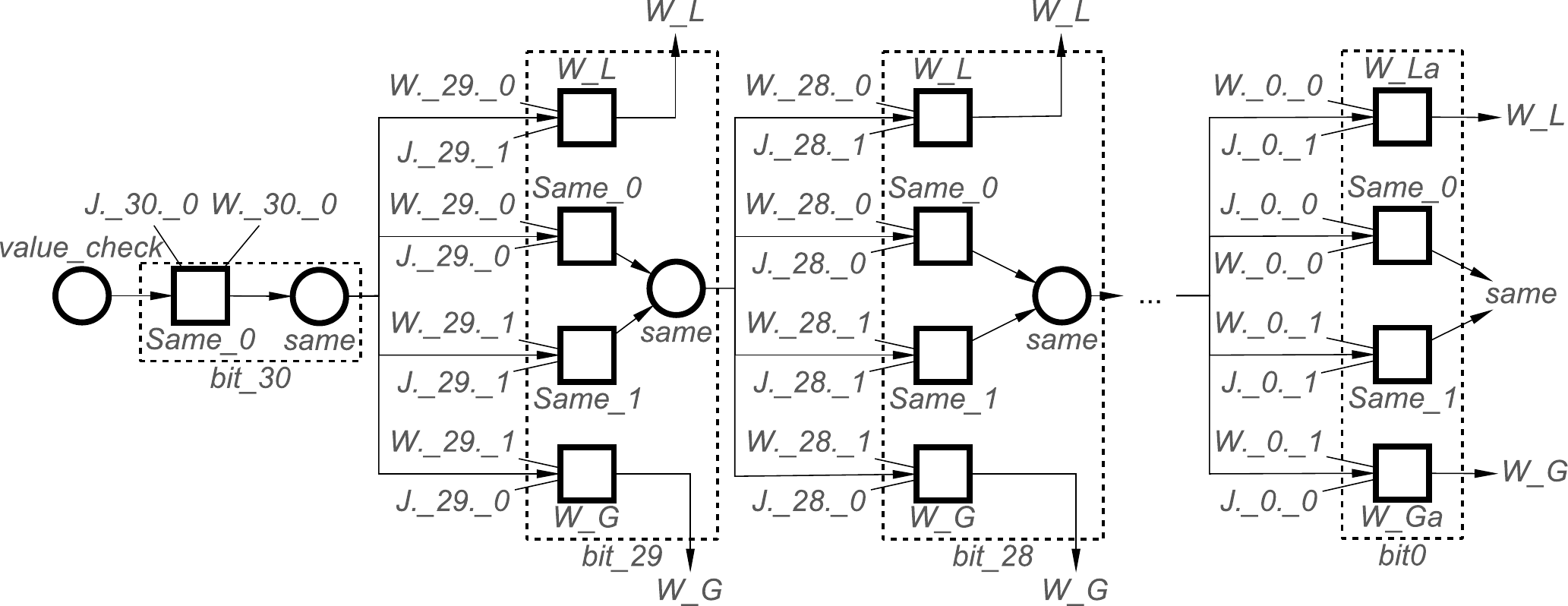} 
        \caption{}
        \label{fig:Signbit:Magnitude}
    \end{subfigure} \\ \vspace{5px}
    \begin{subfigure}[b]{0.18\textwidth}
        \centering
        \includegraphics[width=\linewidth]{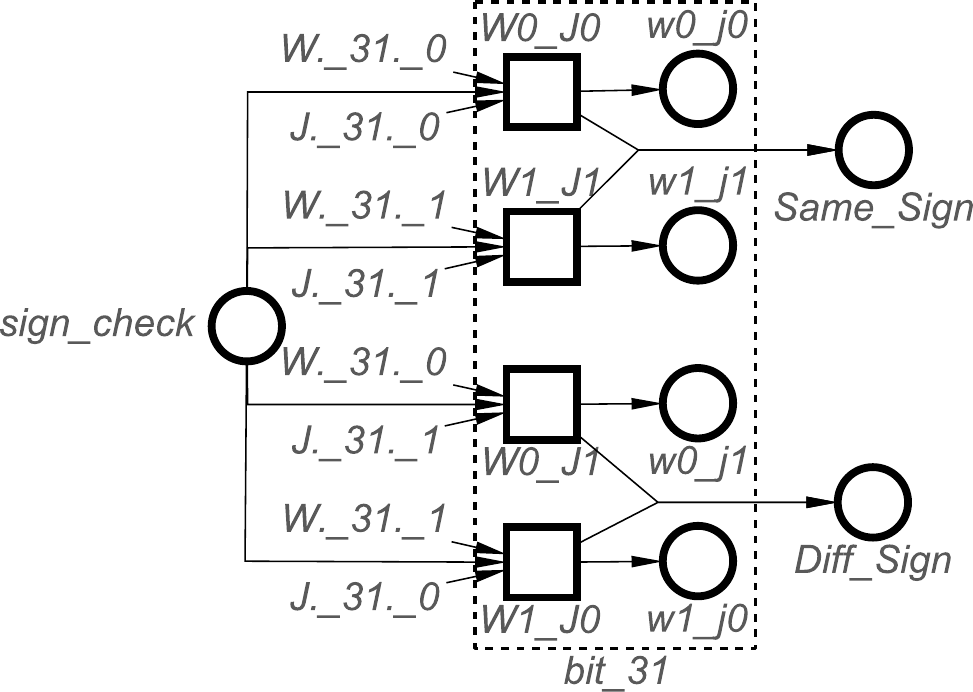}
        \caption{}
        \label{fig:Signbit:SignPair}
    \end{subfigure}
    \begin{subfigure}[b]{0.3\textwidth}
        \centering
        \includegraphics[width=\linewidth]{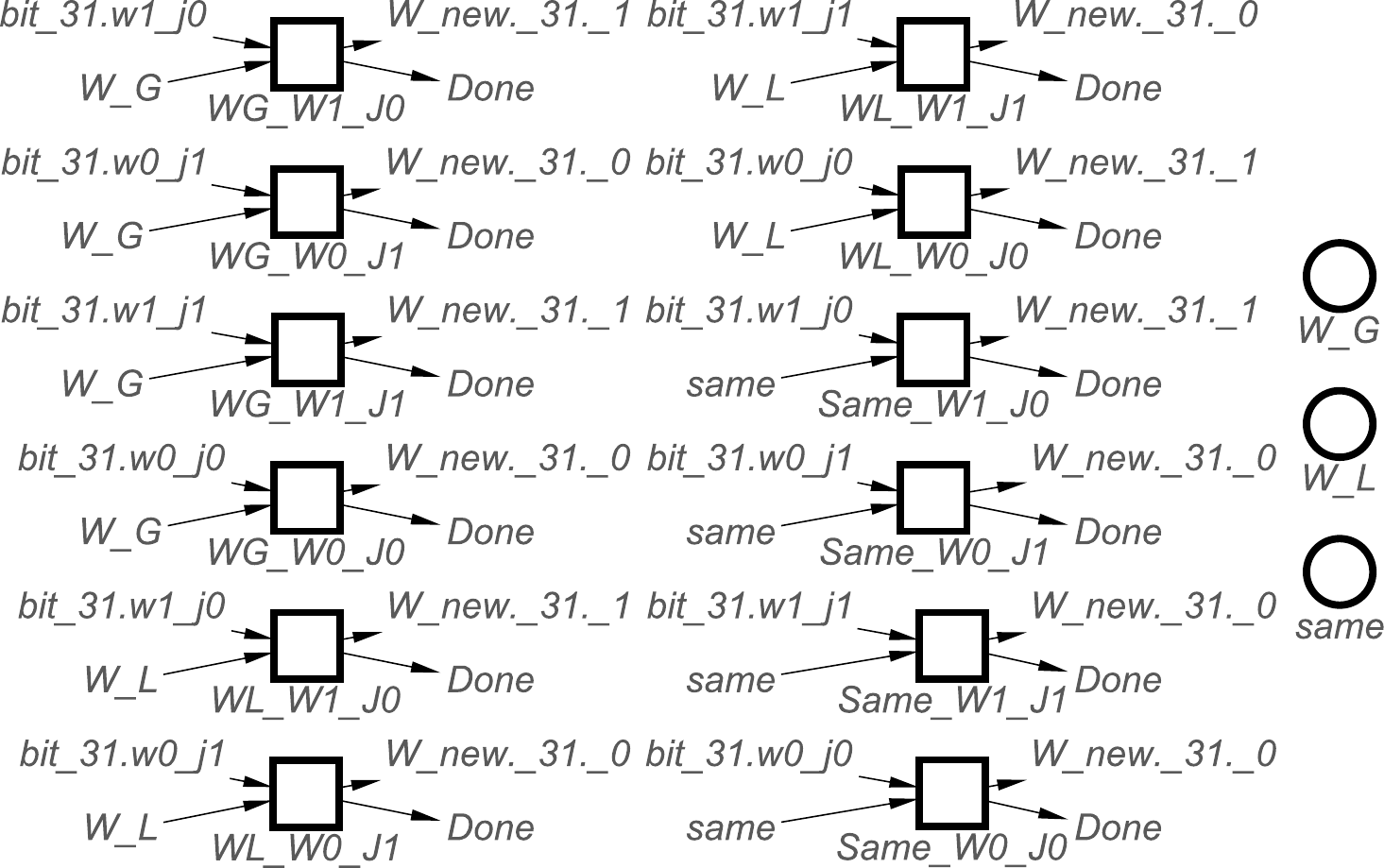}
        \caption{}
        \label{fig:Signbit:NewSign}
    \end{subfigure}
    \caption{PN segment for determining new sign bit. (a) Magnitude comparison of $W$ and $J$. (b) Sign bit pairing of $W$ and $J$. (c)  New sign bit.}
    \label{fig:Signbit}    
\end{figure}

To determine the new sign bit, we first compare the magnitudes of $W$ and $J$. This comparison is carried out bit-by-bit, starting from the most significant bit (bit 30) and proceeding to the least significant bit (bit 0). If $W$ is larger, a token is sent to \textit{W\_G}; if $J$ is larger, a token is sent to \textit{W\_L}; and if the bits are equal, the comparison continues with the next bit; this can be seen in Fig.~\ref{fig:Signbit:Magnitude}. If all bits match through bit 0, a token is sent to Same. Concurrently, we also evaluate the sign-bit pair of $W$ and $J$ (Fig.~\ref{fig:Signbit:SignPair}). Each possible combination of the magnitude result (\textit{W\_G}, \textit{W\_L}, or \textit{Same}) and the sign-bit pair maps to a dedicated transition. The firing of this transition determines the new sign bit for the updated weight (Fig.~\ref{fig:Signbit:NewSign}).

To align the exponents of $W$ and $J$, we shift the mantissa bits of each value according to their respective exponent fields. A positive exponent corresponds to left shifting the mantissa, while a negative exponent corresponds to right-shifting. However, supporting both directions of shifting would greatly increase the complexity of the PN model. To make the design simpler, we restrict ourselves to negative exponents only and therefore implement infrastructure solely for left-shifting. This constraint limits the numerical range of the weights to values between –2 and 2 (exclusive), ensuring that bit 30 is always 0. This is not only compatible with our design but also beneficial, since very large weights can destabilize or saturate a BNN.

\begin{figure}
    \centering
    \includegraphics[width=\linewidth]{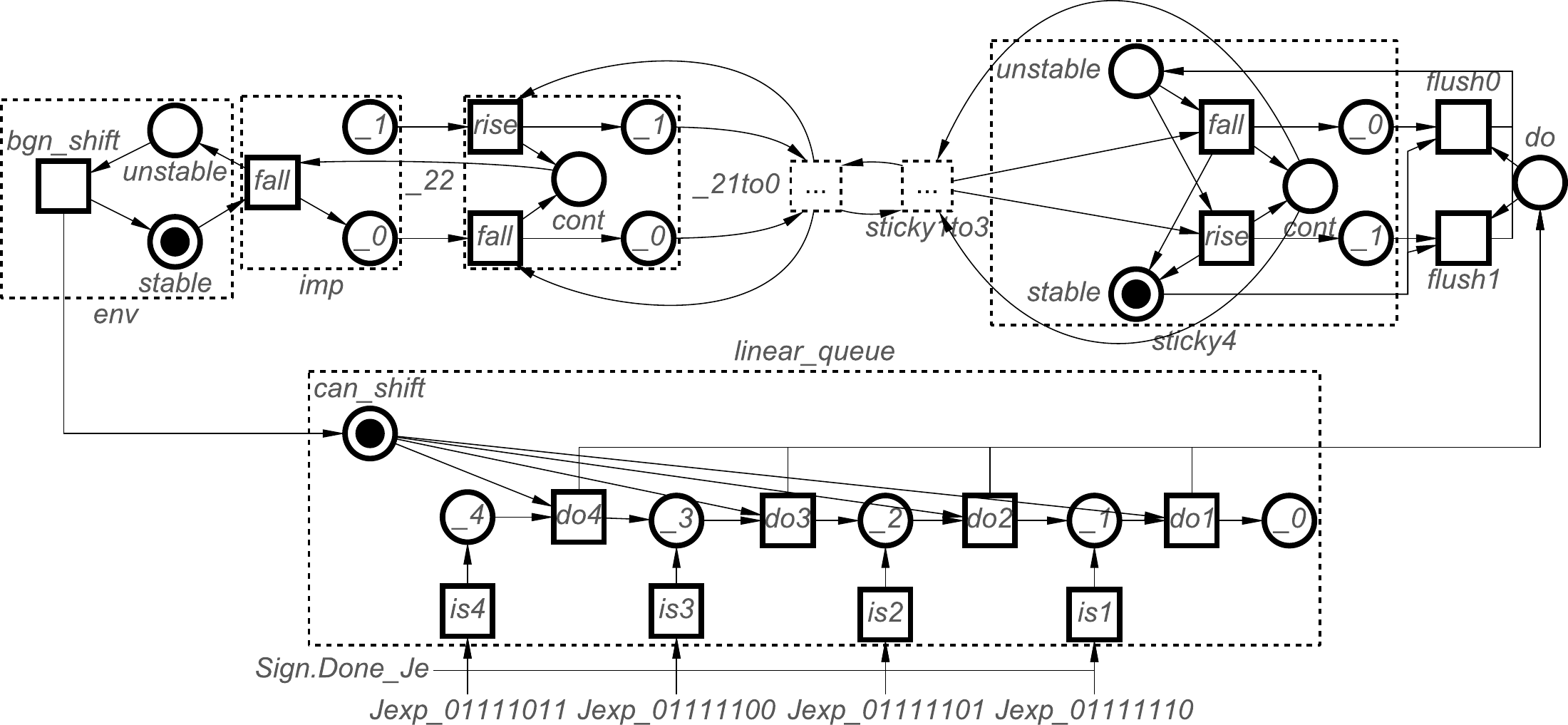}
    \caption{PN segment showcasing infrastructure of how mantissa bits are stored and shifted places to align exponents of $W$ an $J$.}
    \label{fig:ExpAlign}
\end{figure}

Each mantissa is represented in the form 1.Mx23, meaning it includes an implicit leading 1 and a sequence of “sticky bits" after bit 0 to preserve information during repeated shifts. Figure~\ref{fig:ExpAlign} shows a scaled-down example of this shifting mechanism with an implicit-1 block and three value bits. Transitions is\{1, 2, 3, 4\} determine how many shifts are required; each sends a token into the linear queue. The queue then triggers  a transition sending a token to place \textit{do}, which initiates a single left shift of the mantissa. This process repeats until the queue reaches \textit{0}, at which point shifting stops.

For safety, the PN shifts one bit at a time, proceeding from the least significant bit up to the implicit-1 bit. This is enforced by placing a token in a \textit{cont} place for each bit, ensuring that a bit may shift only after the previous bit’s shift (\textit{rise} or \textit{fall}) has completed. Values that shift past bit 0 are discarded, as they no longer contribute to the computation. Note that Fig.~\ref{fig:ExpAlign} is intentionally simplified for clarity. In practice $J$ takes on fixed values and requires at most four left shifts, which means only four sticky bits are needed below bit 0. $W$, by contrast, may be initialized anywhere between magnitude 0 and 2 and may require up to 126 alignment shifts. However, only 24 sticky bits are needed because normalization later limits the useful shift range to at most 24 sticky bits, given the fixed values of $J$.

To decide whether the exponent aligned mantissa bits should be added or subtracted, we examine the sign-bit pairing of $W$ and $J$. The cases are:
\begin{itemize}
    \item If \(W_{\text{Sign}} = J_{\text{Sign}}\): \(W - J = W - J\) or \((-W) - (-J) = -(W - J)\)
    \item If \(W_{\text{Sign}} \neq J_{\text{Sign}}\): \((-W) - J = - (W + J)\) or \(W - (-J) = W + J\)
\end{itemize}

\begin{figure}
    \centering
    \begin{subfigure}[b]{0.475\textwidth}
        \centering
        \includegraphics[width=\linewidth]{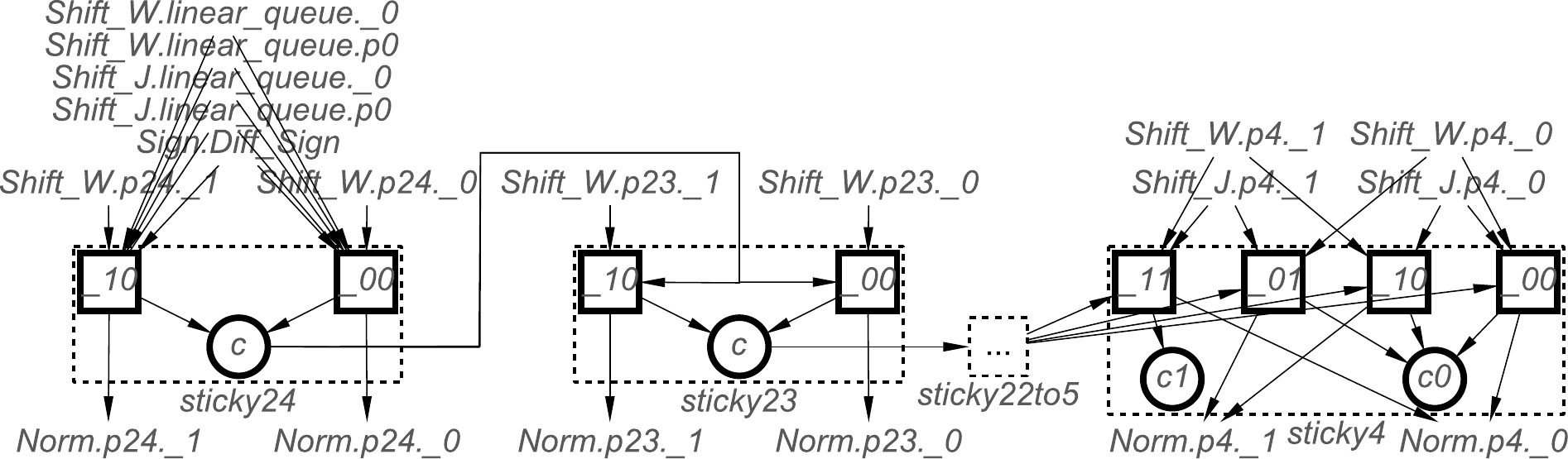} 
        \caption{}
        \label{fig:Mantissa:sticky_for_add_and_sub}
    \end{subfigure} \\ \vspace{5px}
    \begin{subfigure}[b]{0.475\textwidth}
        \centering
        \includegraphics[width=\linewidth]{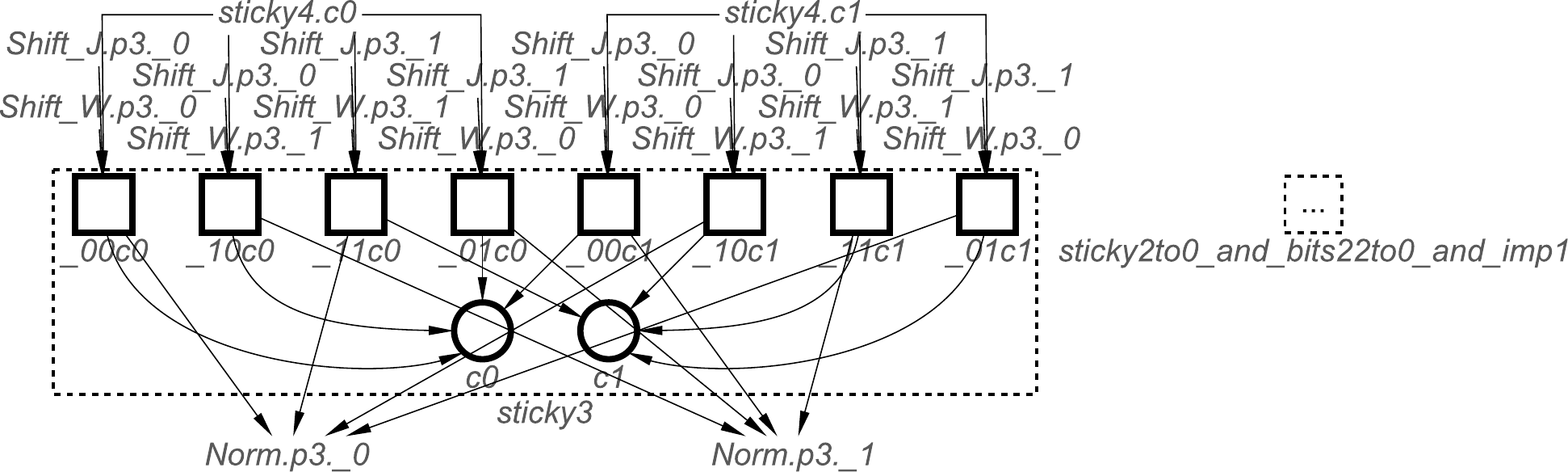} 
        \caption{}
        \label{fig:Mantissa:Add}
    \end{subfigure} \\ \vspace{5px}
    \begin{subfigure}[b]{0.475\textwidth}
        \centering
        \includegraphics[width=\linewidth]{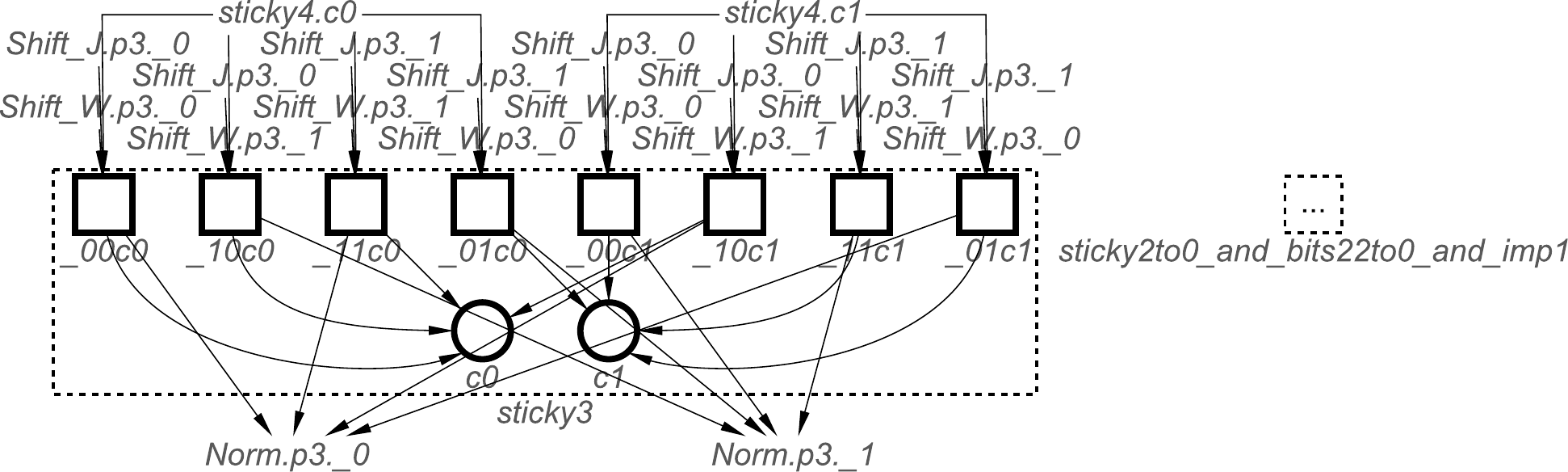}
        \caption{}
        \label{fig:Mantissa:Sub}
    \end{subfigure}
    \caption{PN segment for arithmetic of the mantissa bits: (a) common sticky bits for addition and subtraction operations. (b) Addition of mantissa bits. (c) subtraction of mantissa bits.}
    \label{fig:Mantissa}    
\end{figure}

The addition of mantissa bits is performed bit by bit, from the least significant sticky bit (sticky bit 24) up to the implicit leading 1. At each step, a carry value of 0 or 1 may propagate to the next bit. Fig.~\ref{fig:Mantissa:Add} illustrates the PN segment of the logic for a two-bit example. The subtraction of mantissa bits uses the same bit-by-bit structure but follows the Austrian method of binary subtraction, where values are carried forward rather than borrowing from the next bit. This approach is easier to model in a PN. Fig.~\ref{fig:Mantissa:Sub} shows the PN segment of this for 1 bit but the concept extends to all.

After the mantissa arithmetic has completed, the result must be normalized. Normalization ensures that the implicit 1 leading bit contains a value of 1. To achieve this, the mantissa is repeatedly left-shifted until this condition is met. The procedure mirrors the shifting mechanism in Fig.~\ref{fig:ExpAlign}, but applies shifts in the opposite direction. Unlike exponent alignment, where the exponent value dictates the exact number of shifts, normalization continues until the implicit 1 bit becomes 1. 
Because $W$ may be initialized to any magnitude within its range and $J$ takes on fixed values, we determined that at most 24 shifts are ever required; therefore, 24 sticky bits are sufficient. 
There is one exception: if all mantissa bits are zero, then no shift can produce a leading 1. In this case, a transition \textit{all\_0s} fires and places a token at 0 in the linear queue, indicating that no normalization is needed. The total number of normalization shifts is stored in 8-bit binary form and subtracted from the bias value 127. The result of this subtraction becomes the new exponent for the updated weight.

\begin{figure}
    \centering
    \includegraphics[width=0.8\linewidth]{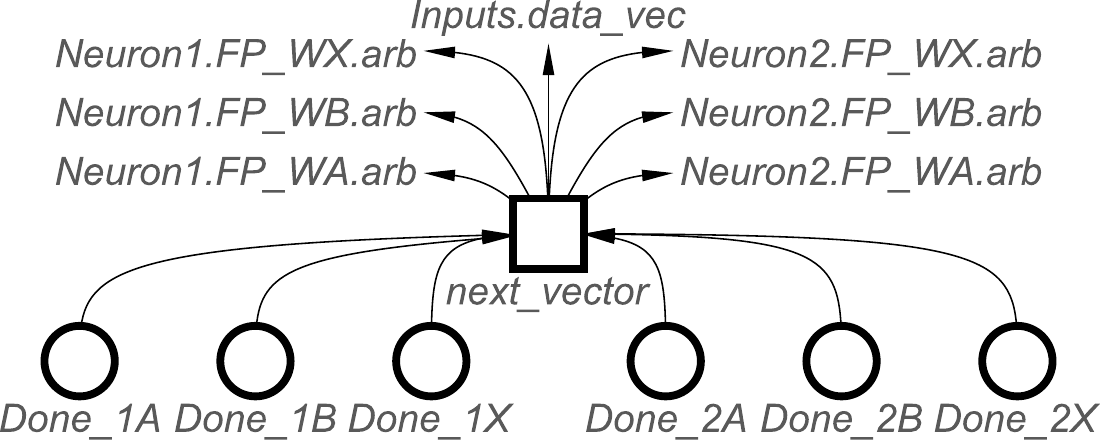}
    \caption{PN segment demonstrating how next data point is loaded once weights have been updated.}
    \label{fig:NextVector}
\end{figure}

The calculation and updating of all six weights in the model occur concurrently. When a weight finishes updating, it produces a token in its corresponding place within Fig.~\ref{fig:NextVector}. After all six places contain a token, the transition \textit{next\_vector} fires, depositing a token into \textit{data\_vec} (Fig.~\ref{fig:Inputs}) to signal that the next data point can be processed. At the same time, a token is sent to \textit{arb} for each weight (Fig.~\ref{fig:WeightBin}), allowing the weights to undergo binarization.

\section{Petri net Verification}\label{sect:verification}
Verification of the BNN PN model is conducted out at three hierarchical levels of abstraction: the segment level (for all segments introduced so far), the component level, and the full system level. At each level, we verify key behavioral and structural properties. These checks are performed using a combination of \reach{} expressions and \workcraft’s \mpsat{} verification tools.
In Table~\ref{tab:verification}, we highlight the properties that we have verified and the method used. More detailed explanation of each verification is discussed in the following sections.
\begin{table}
    \centering
    \caption{Verification checks and methods}
    \begin{tabular}{ll}
    \toprule
         \textbf{Verification Checks} & \textbf{Method of Verification}\\
    \midrule
         Safety/boundedness&\workcraft{} default \\
         Deadlock-freedom&\mpsat{} \\
         Correct token Propagation& Simulation\\
         Hierarchal composition&Mathematical equations\\
         Arbitration& Introduction of arbitration places in segments\\
         Causal sequencing& Simulation\\
         Reversibility& Simulation and design choice\\ \bottomrule
    \end{tabular}
    \label{tab:verification}
\end{table}

\subsection{Structural Verification}
Structural verification checks the correctness of the PNs topology and token-handling guarantees, ensuring that the underlying architecture is well formed.
Safety ensures that no place can hold more than one token at any time ($\forall M \in \mathrm{Reach}(N) : M(p) \le 1$). This prevents ambiguous state configurations and is essential for modeling binary logic, mutual exclusion, and single-control-flow structures. 
Hierarchy preservation verifies that composing multiple segments into larger components does not introduce structural inconsistencies.

\subsection{Behavioral Verification}
Behavioral verification evaluates the dynamic execution of the PN, ensuring design aligns with causal and and computational semantics.
Deadlock-freedom ensures that the PN never reaches a marking in which no transitions are enabled, allowing continuous system progress ($\mathrm{Dead}(M) \equiv \forall t \in T : \neg \mathrm{enabled}(t, M)$).
Reversibility determines whether the system can return to earlier markings, including the initial marking ($\forall M \in \mathrm{Reach}(N, M_0) : M_0 \in \mathrm{Reach}(N, M)$).
Reachability analysis determines whether desired markings are attainable and whether unintended or forbidden markings can occur ($M \in \mathrm{Reach}(N, M_0)$).
Correct token flow and synchronization ensures that the propagation of tokens faithfully captures intended data flow semantics.
Compositional correctness confirms that interactions between segments preserve intended concurrency and ordering.

\subsubsection{Verification Outcomes}
Most individual segments pass verification easily due to their  low complexity design. Verification confirms that operations like Sign, TanH, multiplication, addition, and gradient rules behave as intended. More intricate segments, such as weight binarization, STE computation, and floating-point weight updates, require finer-grained breakdown into sub-segments, which each verifies successfully.

When segments are composed into inference and training components, additional checks ensure correctness of interaction. Here, we verified that the BNN PN confirms to:
\begin{itemize}
    \item Preservation of safety, boundedness, and deadlock-freedom.
    \item Correct propagation of tokens between segments. We can ensure this is inline with the proper simulation and mathematical formulation of the operations modeled (e.g .Sign function segment adheres to equation.~\eqref{eq:Sign}).
    \item Integrity of hierarchical composition.
\end{itemize}

At the full-system level, all structural and behavioral properties are re-evaluated. Key outcomes include:
\begin{itemize}
    \item Arbitration: Weight binarization completes before STE and weight updates, As shown in Fig.~\ref{fig:STE}, an arbitration place receives a token once binarization finishes, enabling the STE computation to proceed by forcing.
    \item Causal sequencing: Neuron output sums precede prediction; prediction precedes loss; loss precedes gradients and updates.
    \item Progressiveness: Model cycles correctly through all four data points, indefinitely. This is assured as \textit{data\_vec} in Fig.~\ref{fig:Inputs} always receives a token when all weights are updated.
    \item Partial irreversibility: System may not return to its initial marking as weight evolution is an inherent characteristic of training.
\end{itemize}
Overall, the model passes all verification checks at every level of abstraction.

\subsection{Significance of Verification}
Formal verification is crucial because PNs represent concurrency, synchronization, and causality. These features provide expressive modeling power but also introduce opportunities for subtle design errors that may not surface through simulation alone. In the context of BNNs: incorrect causal ordering can invalidate computations, synchronization mistakes can distort gradient flow,  unsafe interactions can corrupt weight updates, and misordered control flow can cause deadlocks or unintended markings.

Successful verification provides guarantees that the PN accurately captures the semantics of BNN inference and training, preserves correctness across iterations, and is suitable for reliable validation and future backend code generation.

\section{Validation and Experimental Results}\label{sect:validation-and-experiment-results}
In this section, we validate that the PN model behaves as intended by confirming that it produces the same results as the reference BNN. We compare the outputs of each corresponding operation between the PN model and the reference implementation. We also introduce the design of a PN-based instrumentation model that records key metrics for each data vector during every epoch. In addition, we present an experiment comparing the loss trajectories of the BNN PN and the reference BNN across 100 epochs. Finally, we evaluate the size and scalability of the PN model.

\subsection{Design of Metric Instrument}
\begin{figure} [htbp]
    \centering
    \includegraphics[width=0.9\linewidth]{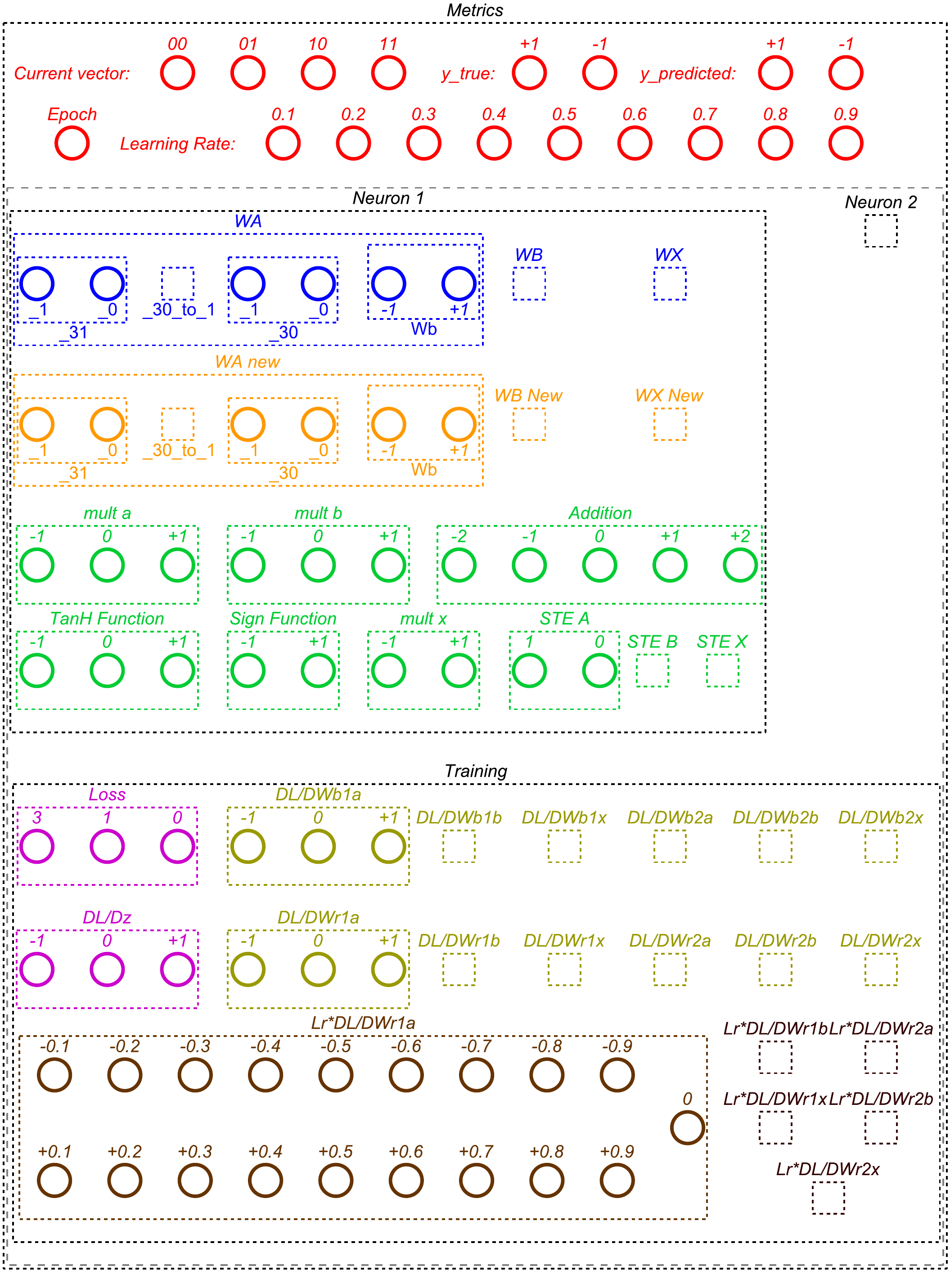}
    \caption{PN model of metric collecting instrument for the BNN PN model showcasing epoch counter, current vector, expected and predicted outputs, learning rate, weights and metrics from all segment calculations.}
    \label{fig:Instrument}
\end{figure}

To enable detailed comparison between the PN model and the reference BNN, we construct a dedicated PN based metric instrument (Fig.~\ref{fig:Instrument}). The instrument records the internal computational values associated with each operation in both the inference and training components. The instrument is organized into four sections:
\begin{enumerate}
    \item Global State Tracking: The top row (RED) includes places that record:
    \begin{itemize}
        \item Current epoch
        \item Active data vector
        \item Expected output ($y_\text{true}$)
        \item Predicted output ($y_\text{predicted}$)
        \item Learning rate
    \end{itemize}
    \item Neuron 1 Metrics
    \begin{itemize}
        \item All bits of the initial real-valued weights (BLUE)
        \item Their binarized form ($W_b$)
        \item All bits of the updated weights (ORANGE)
        \item All inference operation outputs (GREEN)
    \end{itemize}
    \item Neuron 2 Metrics mirrors the Neuron 1 and records corresponding values for the second hidden neuron.
    \item Training Metrics
    \begin{itemize}
        \item Loss and loss derivative values (PINK)
        \item Gradient computations w.r.t each weight (GOLD)
        \item Products of gradients and learning rate (BROWN)
    \end{itemize}
\end{enumerate}

Each transition in the BNN model is connected to its corresponding place in the metric instrument. After all operations for a given data point are completed, but before the next data point is loaded, the tokens stored in these places are flushed to prevent value overlap or unsafe interactions between successive computations. The only exceptions are the learning-rate place, which is initialized once and reused, and the epoch counter, which accumulates one token per completed epoch. 

Overall, the metric instrument is verified to be deadlock-free and safe, with the sole exception of the epoch counter, which is intentionally allowed to hold up to 100 tokens.

\subsection{Experiment Set Up}
To ensure that our PN model executes exactly 100 epochs rather than running indefinitely, we connect a place containing 100 tokens to the first data-point transition (Fig.~\ref{fig:Inputs}). This addition ensures controlled termination but makes the model unsafe. Therefore, it is omitted from diagrams in Section~\ref{sect:Eventizing the Binary Neural Network:Design of Inference Segments}.

We execute the simulation in \workcraft{} and use the PN metric instrument to record metrics for each data point across all epochs. In parallel, we execute the reference BNN for 100 epochs and collect the same metrics. These results are then compared to validate that the BNN PN model behaves as intended. During the \workcraft{} simulation, transitions are fired randomly, which means that the weights and learning rate are initialized randomly. To ensure a fair comparison, we use the values selected by the \workcraft{} simulation when executing the reference BNN.

\subsection{Comparison with Binary Neural Network Behavior}\label{sect:validating_bnn_pn_against_reference}

At a learning rate of 0.6, we compare the running average loss across 100 epochs for both BNN models, as shown in Fig.~\ref{fig:reference_vs_petri_net}.
The PN BNN (red) starts at the same loss level and initially mirrors the reference model’s trajectory. However, around epoch 3, its behavior begins to diverge. This indicates that the PN model performs correct inference at the start, but discrepancies emerge during training—specifically in the weight-update mechanism. After epoch 15, the PN model exhibits a gradual increase in loss, eventually trending toward the reference model’s overall loss behavior. By the end of 100 epochs, the PN model achieves a loss rate roughly 10\% lower than the reference. 

\subsection{Model Size Evaluation}\label{sect:model-size-evaluation}
We provide a breakdown of each PN segment’s structural complexity in terms of the number of places, transitions, arcs, and their total in Table \ref{TableSize}. We also report statistics for a single hidden neuron and for the complete BNN PN model. These measurements help quantify the true scale of the system. 
\begin{table}[htbp]				
\caption{Size of blue print PN segments and BNN model.}
\begin{center}
\begin{tabular}{ccccc}
\toprule
\textbf{Segment} & \textbf{Places} & \textbf{Transitions} & \textbf{Arcs} & \textbf{Total Size} \\
\midrule
 Inputs & 7& 4&   48&59\\
 Weights& 67& 95& 532&694\\
 mult a/b& 5& 4& 12&21\\
 Sum of mult a and b& 5& 9& 27&41\\
 TanH function& 3& 5& 10&18\\
 Sign function& 2& 3& 6&11\\
 mult x& 2& 4& 12&18\\
 STE& 3& 31& 476&510\\
 Hidden Neuron& 238& 409& 3109&3756\\
 Output Sum& 3& 4& 16&23\\
 Prediction& 3& 3& 9&15\\
 Hinge Loss& 10& 12& 42&64\\
 Loss derivative& 21& 3& 21&45\\
 gradient w.r.t $W_\text{bA/B}$& 5& 12& 60&77\\
 gradient w.r.t $W_\text{bX}$& 5& 6& 24&35\\
 gradient w.r.t $W_r$& 3& 6& 18&27\\
 Learning rate& 10& 9& 18&37\\
 LR*gradient& 19& 28& 82&129\\
 Weight update& 1256& 1912& 10642&13810\\
 Next vector& 6& 1& 13&20\\
 Full PN BNN Model& 8243& 12598& 71370&92211\\ \bottomrule\end{tabular}
\label{TableSize}
\end{center}
\end{table}

\input{reference_vs_Petri_net_figure}

Understanding the size of each segment is essential when considering how the model scales to larger architectures with more input, hidden, or output neurons, or with expanded data vectors.
Such insights allow us to extend the model to more complex datasets while preserving explainability, visual clarity, and the ability to validate and verify system behavior.
This exact problem of managing model complexity is a central focus in the formal methods domain, with studies detailing how the bounds on data structure sizes can be applied to maintain verification feasibility~\cite{2025-info_process_letters-drechsler_et_al-lower_bound_proof_for_bdd_sizes}.
Depending on which architectural component is being scaled, some segments must expand in internal size, others must increase in quantity, and many will require both forms of scaling. Further discussion of scalable design strategies will be addressed in future work.

\subsection{Estimated Complexity using Different Datasets}\label{sect:estimated-complexity}
Using Table~\ref{TableSize} as a guideline, we can estimate the approximate size of the BNN PN models of different datasets, such as
6-keyword spotting~(KWS6)~\cite{warden2018speechcommandsdatasetlimitedvocabulary-kws6},
CIFAR2~\cite{Krizhevsky2009_cifar2},
and
MNIST~\cite{binarized_mnist_salakhutdinov2008quantitative}.
To ensure that these sizes reflect real-world implementations, we consider three of the BNN implementations shown in~\cite{2023-ieee_pattern_analysis_and_machine_intelligence-maheshwari_rahman_shafik_yakovlev_rafiev_jiao_granmo-redress_tm} with the respective layers of 128, 256 and 256$\times$128.
Given that the BNN is a single layer and contains two input features and two neurons, the total size of the complete BNN-PN size for a single input feature and a single neuron is around 23053.
We then use this to estimate the model complexity of each dataset's BNN PN by calculating its total size based on each architecture's number of input features, number of hidden layer neurons, and number of output neurons, as shown in Table~\ref{tab:pn_scalability_analysis_for_different_datasets}.
\begin{table}[]
    \centering
    \caption{Estimated complexity of the BNN PN model for different datasets.}
    \begin{tabular}{cccccc}
    \toprule
    \textbf{Dataset}            & \textbf{Neurons}          & \textbf{Places}                   & \textbf{Transitions}          & \textbf{Arcs}     & \textbf{Total Size}   \\
    \textbf{Name}               & \textbf{$\times$ Inputs}  & \textbf{($\times 10^9$)}           & \textbf{($\times 10^9$)}     & \textbf{($\times 10^9$)}         & \textbf{($\times 10^9$)}                      \\ \midrule
    KWS6                    & 134$\times$377    & 0.101 &  0.154 & 0.875 & 1.130 \\
    CIFAR2                  & 130$\times$1024   & 0.270 &  0.414 & 2.344 & 3.028 \\ 
    MNIST                   & 138$\times$784    & 0.210 &  0.320 & 1.813 & 2.343 \\ 
    KWS6                    & 262$\times$377    & 0.202 &  0.309 & 1.750 & 2.261 \\
    CIFAR2                  & 258$\times$1024   & 0.541 &  0.827 & 4.687 & 6.055 \\
    MNIST                   & 266$\times$784    & 0.419 &  0.640 & 3.627 & 4.686 \\
    KWS6                    & 390$\times$377    & 0.268 &  0.410 & 2.320 & 2.998 \\ 
    CIFAR2                  & 386$\times$1024   & 0.608 &  0.930 & 5.267 & 6.805 \\
    MNIST                   & 394$\times$784    & 0.484 &  0.739 & 4.189 & 5.412 \\
    \bottomrule
    \end{tabular}
    \label{tab:pn_scalability_analysis_for_different_datasets}
\end{table}

As the number of neurons increases in the reference BNN designs (or their parts) in Table~\ref{tab:pn_scalability_analysis_for_different_datasets}, the PN model experiences a combinatorial explosion. This is expected in fully instantiated event-based representations, where each feature, weight and neuron contributes to the overall PN size, and therefore highlights the importance of parameter sharing, templating and hierarchical reuse of PN segments to maintain traceability, while preserving the ability to capture the BNN’s causal structure. Fully automating these aspects for any arbitrarily sized dataset remains an open challenge for our future work. 

\section{Conclusion}\label{sect:conclusion}

In this paper, we demonstrated how BNNs can be eventized using PNs, modeling both inference and training  as explicit causal processes. By decomposing the BNN operations into modular PN segments and then composing them into a full system-level model, we expose the causal relationships and dependencies, ordering constraints and concurrency patterns that are typically hidden in numerical implementations. 

Validation against a reference BNN confirming similar behavior, and verification checks performed for PN properties. Size and scalability analysis further demonstrated the structural complexity of this approach when applied to different reference BNN designs or datasets. 

While causal transparency can be achieved, this analysis shows significant model complexity rise, especially in floating-point weight update procedures.
This highlights an obvious tradeoff: causal explainability at the cost of scalability. Nonetheless, this work demonstrates that traditional opaque BNNs can be expressed as analyzable discrete event systems.

This work highlights the potential for applying PNs formal methods to machine learning by offering a pathway from black-box neural computation toward explainable, verifiable and dependable event-driven models. This positions PNs as an effective bridge between ML systems and the correctness guarantees required in safety-critical, resource-constrained, and hardware-driven environments. 

\textit{Future Work:} 
Firstly, we consider incorporating bias terms, which are standard components in NNs like BNNs, and expand the model to support alternative activation functions in the output layer, as well as different loss functions and optimizers.
Next, we consider a dedicated \workcraft{} plugin that can automatically construct these BNN PN models, where the user can provide parameters such as network architecture, output activation functions, loss functions, and optimizers.
Lastly, we consider exploring how the PN models can be represented using a different data structure to speed up current simulation tools of PNs and scale up their causal structure by leveraging the available high-performance computational tools. 

\bibliographystyle{IEEEtran}
\bibliography{references}

\end{document}

%% file: reference_vs_Petri_net_figure.tex
\begin{figure}
    \centering
    \begin{tikzpicture}
      \begin{axis}[
        width=\linewidth,
        height=6cm,
        xlabel={Epochs},
        ylabel={Validation Loss Rate},
        legend pos=south east,
        legend cell align={left},
        grid=both,
      ]
        \addplot[mark=none, color=red] coordinates {
            (1, 0.5)
            (2, 0.375)
            (3, 0.333333333)
            (4, 0.375)
            (5, 0.35)
            (6, 0.333333333)
            (7, 0.321428571)
            (8, 0.3125)
            (9, 0.305555556)
            (10, 0.3)
            (11, 0.295454545)
            (12, 0.291666667)
            (13, 0.288461538)
            (14, 0.291666667)
            (15, 0.305555556)
            (16, 0.307291667)
            (17, 0.308823529)
            (18, 0.310185185)
            (19, 0.311403509)
            (20, 0.3125)
            (21, 0.313492063)
            (22, 0.314393939)
            (23, 0.315217391)
            (24, 0.319444444)
            (25, 0.316666667)
            (26, 0.317307692)
            (27, 0.317901235)
            (28, 0.318452381)
            (29, 0.318965517)
            (30, 0.319444444)
            (31, 0.319892473)
            (32, 0.3203125)
            (33, 0.320707071)
            (34, 0.321078431)
            (35, 0.321428571)
            (36, 0.321759259)
            (37, 0.322072072)
            (38, 0.322368421)
            (39, 0.322649573)
            (40, 0.322916667)
            (41, 0.323170732)
            (42, 0.323412698)
            (43, 0.323643411)
            (44, 0.323863636)
            (45, 0.324074074)
            (46, 0.324275362)
            (47, 0.324468085)
            (48, 0.324652778)
            (49, 0.324829932)
            (50, 0.325)
            (51, 0.325163399)
            (52, 0.325320513)
            (53, 0.325471698)
            (54, 0.325617284)
            (55, 0.325757576)
            (56, 0.325892857)
            (57, 0.326023392)
            (58, 0.326149425)
            (59, 0.326271186)
            (60, 0.326388889)
            (61, 0.326502732)
            (62, 0.329301075)
            (63, 0.332010582)
            (64, 0.334635417)
            (65, 0.337179487)
            (66, 0.339646465)
            (67, 0.342039801)
            (68, 0.344362745)
            (69, 0.346618357)
            (70, 0.348809524)
            (71, 0.350938967)
            (72, 0.353009259)
            (73, 0.355022831)
            (74, 0.356981982)
            (75, 0.358888889)
            (76, 0.360745614)
            (77, 0.362554113)
            (78, 0.364316239)
            (79, 0.366033755)
            (80, 0.367708333)
            (81, 0.369341564)
            (82, 0.370934959)
            (83, 0.37248996)
            (84, 0.374007937)
            (85, 0.375490196)
            (86, 0.376937984)
            (87, 0.37835249)
            (88, 0.379734848)
            (89, 0.381086142)
            (90, 0.382407407)
            (91, 0.383699634)
            (92, 0.384963768)
            (93, 0.386200717)
            (94, 0.387411348)
            (95, 0.388596491)
            (96, 0.389756944)
            (97, 0.390893471)
            (98, 0.392006803)
            (99, 0.393097643)
            (100, 0.394166667)
        };
        \addlegendentry{Petri net};
        \addplot[mark=none, color=blue] coordinates {
            (1, 0.5)
            (2, 0.375)
            (3, 0.416666667)
            (4, 0.4375)
            (5, 0.45)
            (6, 0.458333333)
            (7, 0.464285714)
            (8, 0.46875)
            (9, 0.472222222)
            (10, 0.475)
            (11, 0.477272727)
            (12, 0.479166667)
            (13, 0.480769231)
            (14, 0.482142857)
            (15, 0.483333333)
            (16, 0.484375)
            (17, 0.485294118)
            (18, 0.486111111)
            (19, 0.486842105)
            (20, 0.4875)
            (21, 0.488095238)
            (22, 0.488636364)
            (23, 0.489130435)
            (24, 0.489583333)
            (25, 0.49)
            (26, 0.490384615)
            (27, 0.490740741)
            (28, 0.491071429)
            (29, 0.49137931)
            (30, 0.491666667)
            (31, 0.491935484)
            (32, 0.4921875)
            (33, 0.492424242)
            (34, 0.492647059)
            (35, 0.492857143)
            (36, 0.493055556)
            (37, 0.493243243)
            (38, 0.493421053)
            (39, 0.493589744)
            (40, 0.49375)
            (41, 0.493902439)
            (42, 0.494047619)
            (43, 0.494186047)
            (44, 0.494318182)
            (45, 0.494444444)
            (46, 0.494565217)
            (47, 0.494680851)
            (48, 0.494791667)
            (49, 0.494897959)
            (50, 0.495)
            (51, 0.495098039)
            (52, 0.495192308)
            (53, 0.495283019)
            (54, 0.49537037)
            (55, 0.495454545)
            (56, 0.495535714)
            (57, 0.495614035)
            (58, 0.495689655)
            (59, 0.495762712)
            (60, 0.495833333)
            (61, 0.495901639)
            (62, 0.495967742)
            (63, 0.496031746)
            (64, 0.49609375)
            (65, 0.496153846)
            (66, 0.496212121)
            (67, 0.496268657)
            (68, 0.496323529)
            (69, 0.496376812)
            (70, 0.496428571)
            (71, 0.496478873)
            (72, 0.496527778)
            (73, 0.496575342)
            (74, 0.496621622)
            (75, 0.496666667)
            (76, 0.496710526)
            (77, 0.496753247)
            (78, 0.496794872)
            (79, 0.496835443)
            (80, 0.496875)
            (81, 0.49691358)
            (82, 0.49695122)
            (83, 0.496987952)
            (84, 0.49702381)
            (85, 0.497058824)
            (86, 0.497093023)
            (87, 0.497126437)
            (88, 0.497159091)
            (89, 0.497191011)
            (90, 0.497222222)
            (91, 0.497252747)
            (92, 0.497282609)
            (93, 0.497311828)
            (94, 0.497340426)
            (95, 0.497368421)
            (96, 0.497395833)
            (97, 0.49742268)
            (98, 0.49744898)
            (99, 0.497474747)
            (100, 0.4975)
        };
        \addlegendentry{Reference};
      \end{axis}
    \end{tikzpicture}
    \caption{Graph of running average loss rate for both the reference and Petri net BNNs.}
    \label{fig:reference_vs_petri_net}
\end{figure}
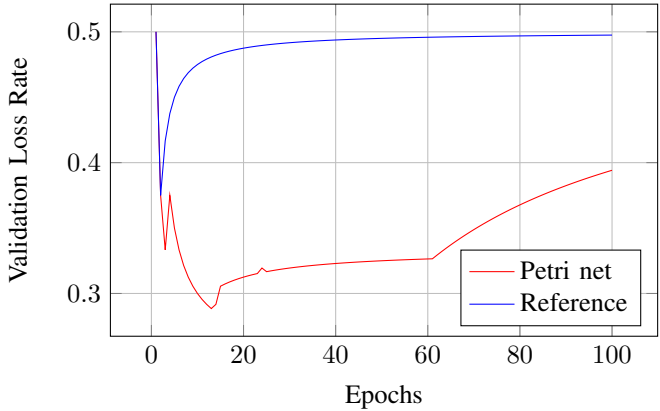